\documentclass[journal]{IEEEtran}
\ifCLASSINFOpdf
   \usepackage[pdftex]{graphicx}
\else
\fi
\usepackage[cmex10]{amsmath}
\usepackage{amssymb}
\usepackage{array}

\def\eop  {{\noindent\framebox[0.5em]{\rule[0.25ex]{0em}{0.75ex}}}}
\def\be {\begin{equation}}
\def\ee {\end{equation}}
\def\beas {\begin{eqnarray*}}
\def\eeas {\end{eqnarray*}}
\def\bea {\begin{eqnarray}}
\def\eea {\end{eqnarray}}
\newtheorem{theorem}{Theorem}
\newtheorem{proposition}{Proposition}
\newtheorem{lemma}{Lemma}
\newtheorem{definition}{Definition}

\newtheorem{claim}{Claim}
\newtheorem{algorithm}{Algorithm}
\newcommand{\cond}{\ |\ }

\newcommand{\argminn}[1]{\underset{#1}{\mathrm{argmin}} \:}
\newcommand{\argmaxx}[1]{\underset{#1}{\mathrm{argmax}} \:}
\newcommand{\argmin}{\mbox{argmin}}
\newcommand{\argmax}{\mbox{argmax}}

\newcommand{\bfone}{\mbox{\bf 1}}
\newcommand{\bfzero}{\mbox{\bf 0}}
\newcommand{\bfa}{{\bf a}}
\newcommand{\bfb}{{\bf b}}
\newcommand{\bfc}{{\bf c}}

\newcommand{\bfp}{{\bf p}}
\newcommand{\bfq}{{\bf q}}

\newcommand{\bfx}{{\bf x}}
\newcommand{\bfy}{{\bf y}}
\newcommand{\bfz}{{\bf z}}
\newcommand{\bai}{\bfb_{\alpha | i}}
\newcommand{\xai}{\bfx_{\alpha} |  x_i}

\newcommand{\bftheta}{\boldsymbol{\theta}}
\newcommand{\bfmu}{\boldsymbol{\mu}}
\newcommand{\bfnu}{\boldsymbol{\nu}}
\newcommand{\bflambda}{\boldsymbol{\lambda}}
\newcommand{\bfsigma}{\boldsymbol{\sigma}}
\DeclareMathOperator{\lp}{\|}

\newcommand{\LG}{{\mathbb L}(G)}

\begin{document}
\title{Norm-Product Belief Propagation: Primal-Dual Message-Passing for Approximate Inference}

\author{Tamir~Hazan~%~\IEEEmembership{Member,~IEEE,}
        %John~Doe,~\IEEEmembership{Fellow,~OSA,}
        and~Amnon~Shashua%,~\IEEEmembership{Life~Fellow,~IEEE}% <-this % stops a space
\thanks{Manuscript resides in arXiv:0903.3127 submitted March 18, 2009 and revised June 16, 2009.}
\thanks{T. Hazan and A. Shashua are with the School of Engineering and Computer Science, Hebrew University of Jerusalem, Jerusalem 91904, Israel 
(e-mail: tamir@cs.huji.ac.il; shashua@cs.huji.ac.il).}% <-this % stops a space
%\thanks{Manuscript received March 19, 2009; revised January 11, 2007.}}
}


%\markboth{Journal of \LaTeX\ Class Files,~Vol.~6, No.~1, January~2007}%
%{Shell \MakeLowercase{\textit{et al.}}: Bare Demo of IEEEtran.cls for Journals}
\markboth{Presented in part at the Conference on Uncertainty in Artificial Intelligence (UAI), July 2008.}%
{Shell \MakeLowercase{\textit{et al.}}: Bare Demo of IEEEtran.cls for Journals}
%\markboth{Technical Report 2009-5, Leibniz Center for Research, School of Computer Science and Engineering, The Hebrew University}%
%{Shell \MakeLowercase{\textit{et al.}}: Bare Demo of IEEEtran.cls for Journals}




\maketitle

\begin{abstract}
%\boldmath
Inference problems in graphical models can be represented as a constrained optimization of a free energy function. In this paper we treat both forms of probabilistic inference, estimating marginal probabilities of the joint distribution and finding the most probable assignment, through a unified message-passing algorithm architecture. In particular we generalize the Belief Propagation (BP) algorithms of sum-product and max-product and tree-rewaighted (TRW) sum and max product algorithms (TRBP) and introduce a new set of convergent algorithms based on "convex-free-energy" and Linear-Programming (LP) relaxation as a zero-temprature of a convex-free-energy. The main idea of this work arises from taking  a general perspective on the existing BP and TRBP algorithms while observing that they all are reductions from the basic optimization formula of $f + \sum_i h_i$ where the function $f$ is an extended-valued, strictly convex but non-smooth and the functions $h_i$ are extended-valued functions (not necessarily convex). We use tools from convex duality to present the "primal-dual ascent" algorithm which is an extension of the Bregman successive projection scheme and is designed to handle optimization of the general type $f + \sum_i h_i$. We then map the fractional-free-energy variational principle for approximate inference onto the optimization formula above  and introduce the "norm-product" message-passing algorithm. Special cases of the norm-product include sum-product and max-product (BP algorithms), TRBP and NMPLP algorithms. When the fractional-free-energy is set to be convex (convex-free-energy) the norm-product is globally convergent for the estimation of marginal probabilities and for approximating the LP-relaxation. We also introduce another branch of the norm-product which arises as the "zero-temerature" of the convex-free-energy  which we refer to as the "convex-max-product". The convex-max-product is convergent  (unlike max-product) and aims at solving the LP-relaxation.
\end{abstract}


% \fi
\IEEEpeerreviewmaketitle

\section{Introduction}
\IEEEPARstart{P}{robabisitic} graphical models present a convenient and popular tool for reasoning about complex distributions. The graphical model reflects the way the complex distribution $p(x_1,...,x_n)$ factors into a product of  potential functions, each defined over a small number of variables, and referred to as {\it factors}. A graphical model, which defined in terms of {\it factor graphs\/}, represents the incidence between factors and the variables by a bipartite graph with one set of nodes corresponding to the variables of the joint distribution and another set of nodes standing for the factors. An edge exists between a variable node and a factor node if the variable is contained in the set of variables represented by the factor. In many applications of interest the factor graph is sparse. In other words, in the modeling of the joint behavior of a set of interacting variables it is often the case that only a small subset of variables interact directly.
For example, in the domain of image processing, if we think of each pixel as a variable in a joint distribution over all image pixels then, typically the intensity value of a single pixel will depend most strongly on neighboring pixels in the image, rather than on those at a distant location. Without the local interaction assumption, i.e., if each variable interacts directly with all other variables, then the inference of the joint behavior would be a hopeless task. 

Problems involving {\it inference\/} using graphical models comes up in a wide range of applications covering a variety of disciplines. Those include digital communications (error correcting codes \cite{Feldman-allerton03}), computer vision \cite{Tappen-iccv03}, medical diagnosis \cite{Jaakkola-jair99}, protein folding \cite{Yanover-nips03}, computer graphics \cite{Freeman-ijcv00,Avidan-cvpr08}, clustering \cite{Shental-iccv03}, as well as other broad disciplines which include signal processing, artificial intelligence and statistical physics \cite{Frey98,Jordan-book98}.

Probabilistic inference comes in two distinct forms and typically involve two slightly different algorithmic thrusts. One form of inference task is to obtain one global state of the joint distribution that is most probable, i.e., find the values of $x_1,...,x_n$ which maximizes $p(x_1,...,x_n)$. This form of inference is typically referred to as the  {\it maximal a-posteriori\/} assignment, or in its abbreviated form, the MAP assignment. The second type of inference has the objective of obtaining marginal probabilities for some subset of variables given evidence (value of) about other variables. For example, if $x_i\in\{1,...,n_i\}$ then $p(x_i)$ comes out of summing exponentially many elements $\sum_{\{x_1,...,x_n\}\setminus x_i} p(x_1,...,x_n)$ resulting in the likelihood of $ x_i$ to obtain each of its possible $n_i$ values. In this paper, we will focus on {\sl both\/} inference problems with the objective of introducing a unifying algorithmic thrust.

Exact inference is NP-hard \cite{Shimony-ai94}, thus introducing the need to derive algorithms for {\it approximate inference}. One of the most popular class of methods for inference over (factor) graphs are message-passing algorithms which pass messages along the edges of the factor graph until convergence is reached. The belief-propagation (BP) algorithms \cite{Pearl-88} come in two variations: the {\it sum-product\/} algorithm for computing marginal probabilities and the {\it max-product\/} algorithm for computing the MAP assignment. Citing \cite{Yedidia05}, the centrality of inference using graphical models and the utility of the  BP algorithms for solving them is reflected in the fact that equivalent or very similar message-passing algorithms have been independently derived under different disciplines. Those include the Viterbi algorithm \cite{Viterbi67}, Gallager's sum-product algorithm for decoding low-density parity check codes \cite{Gallager63}, the turbo-decoding algorithm \cite{turbo-codes}, the Kalman filter for signal processing \cite{Kalman60}, and the transfer-matrix approach in statistical mechanics \cite{Baxter82}.

The BP algorithms are exact, i.e., the resulting marginal probabilities and the MAP assignments are the correct ones, when the factor graph is free of cycles --- a state of affairs that considerably limits the application of those algorithms to solve real world problems. Nevertheless, 
an intriguing feature of BP, which most likely is the source for its great popularity, is that it is well-defined and often gives surprisingly good approximate results for graphical models with cycles. However, in this context  there are no convergence guarantees (except under some special cases \cite{Tatikonda02}, \cite{Mooij-Kappen}) and the algorithms fail to converge in many cases of interest.

During the past  decade there has been much progress in putting forward a framework for approximate inference using variational principles. It has been shown that the fixed-points of the sum-product algorithm (for estimating marginal probabilities) correspond to the fixed-points of a constrained energy function called the Bethe free energy \cite{Yedidia05}. The free energy arises from the expansion of the KL-divergence between the input distribution and its product form. The Bethe approximation replaces the entropy term in the free energy by the Bethe entropy. The investigation of the stationary points of the Bethe free energy yields conditions for convergence of BP \cite{Heskes04}, and lower bounds for the free energy in some special cases \cite{Sudderth08}. These lower bounds are based on the loop calculus framework which considers the Bethe free energy as a first order approximation for the free energy \cite{Chertkov06}. The Bethe free energy is exact for factor graphs without cycles, as well as convex over the set of constraints (representing validity of marginals). When the factor graph has cycles the Bethe energy is non-convex and the BP algorithms  may fail to converge. Although it is possible to derive convergent algorithms to a local minima of the Bethe function \cite{yuille02}, \cite{Heskes06} the computational cost is large and thus has not gained popularity. 

To overcome the difficulty with the non-convexity of the Bethe approximation, several authors have introduced a class of approximations known as {\it convex free energies\/} which are convex over the set of constraints for any factor graph. An important member of this class is the tree-reweighted (TRW) free energy which consists of a linear combination of free energies  defined on spanning trees of the factor graph \cite{Wainwright-nips02}. It is notable that for this specific member of convex free energies a convergent message-passing algorithm, applicable to pairwise factors only, has been recently introduced \cite{Globerson-UAI07}.  However, 
a convergent message passing algorithm for the general class of convex free energies is still lacking. The existing algorithms either employ damping heuristics  to ensure convergence in practice \cite{Wainwright-05upper} or focus on a sub-class of free energies where the entropy term is a positive combination of joint entropies \cite{Heskes06}. 

The MAP assignment problem has been shown to be approximated by a Linear-Programming (LP) relaxation scheme \cite{Wainwright-05map} with message-passing algorithmic attempts as a solution \cite{Kolmogorov-uai05, Weiss-uai07,Globerson-nips07, Werner07, Meltzer-uai09}. Some of these attempts guarantee convergence only under special cases (such as binary variables), \cite{Kolmogorov-uai05, Weiss-uai07}. Others, such as \cite{Globerson-nips07}, arises as a special case of our algorithm. We refer to \cite{Meltzer-uai09} for detailed account on the connections between these message-passing algorithms. A double-loop of message passing using a proximal minimization technique proposed recently by \cite{Ravikumar-etal-icml08} is convergent but at a considerable computational expense. Dual decomposition techniques were recently proposed \cite{Kolmogorov-pami06, Komodakis-pami10}, which are related to dual subgradient methods for the LP relaxation.

In this paper, we derive a class of approximate inference message-passing algorithms, which we call {\it norm-product\/} algorithms, using the notion of free-energy approximation. The norm-product is an inference engine covering both the estimation of marginal probabilities and the MAP assignment. When the Bethe free energy is used as a substitution for the free-energy, the norm-product reduces to the sum-product and max-product algorithms where the latter emerges as a "zero temperature" version of the former. When a convex-free-energy is used the norm-product becomes a convergent family of algorithms along three strains: (i) a globally convergent algorithm, which we call {\it convex-sum-product\/}, for estimating marginal probabilities, (ii) a locally convergent algorithm emerging as a zero-temperature version of the former strain, we call {\it convex-max-product\/}, for estimating the MAP assignment, and (iii) a globally convergent algorithm for the LP-relaxation problem.

The convex-sum-product algorithm was published in \cite{Hazan-Shashua-uai08} with only a brief sketch of the detailed derivation.  In this paper we have chosen to put a large amount of material in appendices. Due to the complexity of the presented material and the extensive use of modern optimization infrastructure, the body of the paper contains the main "storyline", statements and algorithms whereas the detailed proofs and the required mathematical infrastructure are contained in appendices.

\section{Notations, Problem Setup and Background}

Let $x_1,...,x_n$ be the realizations of $n$ discrete random variables where the range of the $i'th$ random variable is $\{1,...,n_i\}$, i.e., $x_i\in\{1,...,n_i\}$. We consider a joint distribution $p(x_1,...,x_n)$ and assume that it factors into a product of non-negative functions (potentials):
\be 
\label{eq:fact}
p(x_1,...,x_n)=\frac{1}{Z}\prod_{i=1}^n \phi_i(x_i)\prod_{\alpha=1}^m \psi_\alpha (\bfx_\alpha),
\ee
where the functions $\phi_i(x_i)$ represent "local evidence" or prior data on the states of $x_i$, and the functions $\psi_\alpha (\bfx_\alpha)$ have arguments $\bfx_\alpha$ that are some subset of $\{x_1,...,x_n\}$ and $Z$ is a normalization constant, typically referred as the {\em partition function}. For example, $p(x_1,x_2,x_3)=(1/Z)\psi_{23}(x_2,x_3)\psi_{13}(x_1,x_3)$ has two factors with indices $\alpha_1=\{2,3\},\alpha_2=\{1,3\}$ and $\bfx_{\alpha_1}=\{x_2,x_3\}, \bfx_{\alpha_2}=\{x_1,x_3\}$, and uniform local evidence $\phi_i(x_i)=1$ for every $i=1,2,3$ and every $x_i$. 

The factorization structure above defines a {\it hypergraph\/} whose nodes represent the $n$ random variables and the subsets of variables $\bfx_\alpha$ correspond to its hyperedges. For example, if all factor functions are defined on pairs of random variables then the factorization is represented by a graph. A convenient way to represent hypergraphs is by a bipartite graph with one set of nodes corresponding to the original nodes of the hypergraph and the other set corresponds to its hyperedges. In the context of graphical models such a bipartite graph representation is referred to as 
 a {\it factor graph\/} \cite{Kschischang-Frey01}  with {\it variable nodes\/} representing $\phi_i(x_i)$  and a {\it factor node\/} for each function $\psi_\alpha(\bfx_\alpha)$. An edge connects a variable node $i$ with factor node $\alpha$ if and only if $ x_i \in \bfx_\alpha$, i.e., $ x_i$ is an argument of $\psi_\alpha$. %(see Fig.~\ref{fig:factor-graph} for illustration). 
 We adopt the terminology where $N(i)$ stands for all factor nodes that are neighbors of variable node $i$, i.e., all the nodes $\alpha$ for which $ x_i \in \bfx_\alpha$, and $N(\alpha)$ stands for all variable nodes that are neighbors of factor node $\alpha$.
 
We shall focus on the two inference tasks of computing marginal probabilities and maximum a-priori (MAP) assignment. The computation of the marginal probabilities $p(x_i)=\sum_{\bfx\setminus x_i} p(\bfx)$ and $p(\bfx_\alpha)=\sum_{\bfx\setminus \bfx_\alpha} p(\bfx)$, requires the summation over the states of all the variable nodes not in $x_i$ or $\bfx_\alpha$ respectively. This computation is generally hard because it may require summing up exponentially large number of terms --- thus one seeks efficient ways or approximate solutions for the marginals. The MAP assignment is the task of finding a state for each $ x_i$ that brings the maximal value to the joint probability $p(x_1,...,x_n)$. 
 
 The belief-propagation (BP) algorithms, known as sum-product and max-product, are two algorithms for computing marginal probability and MAP assignment, respectively, that can be described in terms of operations on a factor graph. As already mentioned in the introduction, the BP algorithms will deliver the correct inference, i.e. are exact, if the factor graph has no cycles, but are still well defined and often provide good approximate results when the factor graph has cycles.
 
 The BP algorithms are defined in terms of {\it messages\/} between variable and factor nodes. The message $m_{\alpha\rightarrow i}(x_i)$ from factor node $\alpha$ to variable node $i$, and the opposite direction message $n_{i\rightarrow\alpha}(x_i)$, is a vector over the states of $ x_i$. In the sum-product algorithm those have the following form:
 \beas
m_{\alpha\rightarrow i}(x_i)&=&\sum_{ \bfx_\alpha\setminus x_i}\psi_\alpha(\bfx_\alpha)\prod_{j\in N(\alpha)\setminus i}n_{j\rightarrow\alpha}(x_j)\\
n_{i\rightarrow\alpha}(x_i)&\propto&  \phi_i(x_i) \prod_{\beta\in N(i)\setminus \alpha}m_{\beta\rightarrow i}(x_i)
\eeas
The $\propto$ indicates that one can normalize the vector.  The messages $n_{i\rightarrow\alpha}(x_i)$ are usually initialized to the uniform vector. Upon convergence of the message-passing scheme the marginal probabilities $p(x_i)$ and $p(\bfx_\alpha)$ can be expressed in terms of a "pseudo-distribution", also known as {\it beliefs}, $b_i(x_i)$ and $b_\alpha(\bfx_\alpha)$ defined below:
\beas
b_i(x_i)&\propto& \phi_i(x_i) \prod_{\alpha\in N(i)}m_{\alpha\rightarrow i}(x_i)\\
b_\alpha(\bfx_\alpha)&\propto&  \psi_\alpha(\bfx_\alpha)\prod_{j\in N(\alpha)}n_{j\rightarrow\alpha}(x_j)
\eeas
When the factor graph has no cycles the messages converge and the beliefs correspond to the marginal probabilities. When the factor graph has cycles there is no convergence guarantee and, regardless of convergence, the recovered beliefs provide only an approximation to the marginal probabilities.

In the max-product algorithm the messages $m_{\alpha\rightarrow i}(x_i)$ are slightly altered:
$$m_{\alpha\rightarrow i}(x_i)=\max_{ \bfx_\alpha\setminus x_i}\left\{\psi_\alpha(\bfx_\alpha)\prod_{j\in N(\alpha)\setminus i}n_{j\rightarrow\alpha}(x_j)\right\},$$
while $n_{i\rightarrow\alpha}(x_i)$ remain as in the sum-product algorithm. The MAP assignment can be recovered from the beliefs $b_i(x_i)$ when the factor graph is a tree. In such a case, the MAP assignment of $ x_i$ corresponds to the index of highest entry of $b_i(x_i)$. In general convergence is not guaranteed, and the MAP assignment can be recovered only for specific problems, \cite{Weiss00, Bayati05, Huang07, Sanghavi07}.   

\subsection{Inference using a Variational Principle}
\label{sec:var-inference}

The BP algorithms apply to tree-structured factor graphs  yet are well defined for general factor graphs but without convergence or accuracy guarantees. The variational principle approach, described below, is a decade long effort at providing an extended platform from which old, i.e., BP algorithms, and new (preferably convergent) algorithms can emerge.

The variational approach seeks a distribution $p(x_1,...,x_n)$ that is as close as possible, in relative entropy terms, to the product potentials $\phi_i(x_i)$ and  $\psi_\alpha(\bfx_\alpha)$. Expanding the KL-divergence $D(\bfp\ ||\ \bfq)=\sum_x p(x)\ln (p(x)/q(x))$ between two nonnegative vectors results in:
$$D(\bfp \ ||\ \prod_i \phi_i \prod_\alpha \psi_\alpha) = F(\bfp),$$
where $F(\bfp)$ is the so called Gibbs-Helmholtz free-energy:
\bea
F(\bfp) &=& \sum_{i,x_i} \theta_i(x_i) p(x_i) + \sum_{\alpha,\bfx_\alpha} \theta_\alpha(\bfx_\alpha) p(\bfx_\alpha) - H(\bfp)
\nonumber\\
&=& E(\bfp) - H(\bfp)
\label{eq:free-energy}
\eea
The term $H(\bfp) = - \sum_{\bfx}p(\bfx)\ln p(\bfx)$ is the entropy and $\theta_i = -\ln \phi_i$ and $\theta_\alpha = -\ln \psi_\alpha$. The linear term $E(\bfp)$ is often referred to as the {\it energy\/} term.
By minimizing $F(\bfp)$ over the probability simplex ${\cal P}=\{\bfp :\ \bfp \ge 0, \sum_{\bfx}  p(\bfx) =1\}$ we get back the probability distribution defined in eqn.~\ref{eq:fact}, as the optimal argument $\bfp^* = \mbox{argmin}_{\bfp\in {\cal P}} F(\bfp)$, and minus the log of the normalization, or equivalently the partition function, as the value at the minimum:
$$\bfp^* = \frac{1}{Z}\prod_{i=1}^n \phi_i(x_i)\prod_{\alpha=1}^m \psi_\alpha (\bfx_\alpha),\ \ \ \ -\ln Z = F(\bfp^*).$$
Since $F(\bfp)$ is strictly convex and the simplex constraints are convex, the minimum is unique. So far we have not gained anything because the entropy $H(\bfp)$ is computationally intractable since its evaluation is exponential in $n$, and satisfying the probability simplex constraints is intractable as well. The variational methods are based on a tractable approximation to the free-energy $F(\bfp)$ by (i) approximating the entropy term $H(\bfp)$ by a combination of local entropies over marginal probabilities $p(x_i),p(\bfx_\alpha)$, and (ii) by approximating the probability simplex constraints by the so called "marginal consistency" constraints.

In approximate inference, the true marginal distributions $p(x_i)$ and $p(\bfx_\alpha)$ are replaced by "beliefs" $b_i(x_i)$ and $b_\alpha(\bfx_\alpha)$ which form a "pseudo distribution" in the sense that 
the beliefs might not necessarily arise as marginals of some distribution $p(x_1,...,x_n)$. The probability simplex constraints are replaced by {\it marginal consistency\/} constraints $\LG$ defined below:

{\small
$$\LG = \left\{ \bfb=\{\bfb_i,\bfb_\alpha\} :
\begin{array}{l} \displaystyle \sum_{\bfx_\alpha\setminus x_i}b_\alpha(\bfx_\alpha) = b_i(x_i) \;\;\; \forall i, \alpha\in N(i) \\
b_\alpha(\bfx_\alpha) \ge 0, \; \sum_{\bfx_\alpha}b_\alpha(\bfx_\alpha) = 1 \;\;\;  \forall \alpha \\
\end{array}
\right.
$$
}
The entropy approximation $\tilde H(b)$ as a function of the beliefs  is known as {\it fractional entropy\/} and has the form:
\be  \sum_\alpha \bar c_\alpha H(\bfb_\alpha) + \sum_i \bar c_i H(\bfb_i),\label{eq:Hcon}\ee
where the joint entropy $H(\bfb_\alpha)=-\sum_{\bfx_\alpha} b_\alpha(\bfx_\alpha)\ln b_\alpha(\bfx_\alpha)$ and the local entropy $H(\bfb_i)=-\sum_{x_i}b(x_i)\ln b(x_i)$.

For factor-graphs without cycles, the setting of $\bar c_\alpha =1$ and $\bar c_i=1-d_i$ where $d_i$ is the degree of the variable node associated with $ x_i$ in the factor graph, renders the approximation $\tilde H$ to be {\it exact\/} and equal\footnote{in this case the joint probability can be expressed solely in terms of the marginals: $p(x_1,...,x_n)=\prod_\alpha p(\bfx_\alpha)/\prod_i p(x_i)^{d_i-1}$. Expanding $H(p)$ produces the Bethe entropy approximation.}
to the entropy $H$. Such an approximation is known as the {\it Bethe\/} entropy:
$$H_{bethe}(\bfb)\stackrel{def}{=}\sum_\alpha H(\bfb_\alpha) + \sum_i (1 - d_i)H(\bfb_i).$$
Moreover, in the case of a tree, the marginal consistency constraints $\LG$ are equal to the probability simplex constraints, thus making the constrained {\it Bethe free energy\/} problem
$$\min_{\bfb \in \LG} \sum_{i,x_i} \theta_i(x_i)b_i(x_\alpha) + \sum_{\alpha,\bfx_\alpha} \theta_\alpha(\bfx_\alpha)b_\alpha(\bfx_\alpha) - H_{bethe}(\bfb),$$
a convex optimization producing the true marginals $b_i(x_i)=p(x_i)$ and $b_\alpha(\bfx_\alpha)=p(\bfx_\alpha)$. The constrained optimization is 
defined in terms of beliefs only and is therefore computationally tractable. However, if the factor graph has cycles, the minimizer of the constrained Bethe free energy is not guaranteed to correspond to the true marginals $p(x_i)$, $p(\bfx_\alpha)$, and not even realizable as a true distribution. Therefore, for general factor graphs, the Bethe free energy optimization approach finds an approximation to the true marginal probabilities. From the optimization point of view, the Bethe free energy is strictly convex in the intersection of constraints when the factor graph is a tree.
When the factor graph has cycles the Bethe energy is non-convex and although it is possible to derive convergent algorithms to local minima of the Bethe function \cite{yuille02}, \cite{Heskes06} the computational cost is large and thus has not gained popularity.

What makes the Bethe free energy optimization interesting is the observation, first elucidated by \cite{Yedidia05}, that when the sum-product algorithm  converges then it does so to a stationary point of the constrained Bethe free energy, i.e., fixed-points of the algorithm correspond to stationary points of the variational problem. This does not mean that the sum-product algorithm descends on the Bethe free energy (in fact it does not), but that near a fixed point things start to behave well. The significance of the observation is that it ties the popular sum-product algorithm with a specific variational principle and moreover it suggests a framework for seeking natural generalizations of the Bethe approximation with their associated message-passing algorithms.

Generalizations of the Bethe free energy move along two thrusts. The first employs better (higher-order) approximations to the entropy and higher-order constraints beyond the marginal consistency constraints to better approximate the full probability simplex constraints. This effort includes Kikuchi free energy, region graphs and other hyper-graph based methods \cite{Yedidia05,Kikuchi51}. The second thrust looks for convergence guaranteed message-passing algorithms by extending the Bethe free energy to form a wider class of functions, known as {\it convex free energies}, which are convex in the intersection of marginal consistency constraints. In this paper we focus on the second thrust. The inclusion of Kikuchi approximations and region graphs is a natural extension to the results we introduce in this paper but for the sake of clarity we leave it outside the current scope.

The fractional entropy eqn.~\ref{eq:Hcon} can be set to form a family of concave approximations. The set of sufficient conditions for an entropy approximation of the type of eqn.~\ref{eq:Hcon} to be concave over the set of constraints was introduced in \cite{Heskes-nc04,Weiss-uai07} and take the following form:\\
\begin{definition}[Concave Entropy Approximation]
\label{def:cfe}
An approximate entropy term of the form eqn.~\ref{eq:Hcon} is strictly concave over the set of marginal consistency constraints  if there exists $c_i,c_{i \alpha} \ge 0$ and $c_{\alpha} > 0$ such that $\bar c_{\alpha} = c_{\alpha} + \sum_{i \in N({\alpha})} c_{i \alpha}$ and $\bar c_i = c_i - \sum_{\alpha \in N(i)} c_{i \alpha}$. The approximate entropy $\tilde H(\bfb)$ becomes:

{\small
\be
\sum_\alpha c_\alpha H(\bfb_\alpha) + \sum_i c_i H(\bfb_i) + \sum_{i,\alpha\in N(i)} c_{i \alpha} (H(\bfb_\alpha) - H(\bfb_i)).\label{eq:H}
\ee}
\end{definition}
The entropy approximation $\tilde H(\bfb)$ includes the Bethe approximation when $c_\alpha=1, c_i=1-d_i$ and $c_{i\alpha}=0$ but it is guaranteed to be strictly concave for any setting of the parameters where $c_i,c_{i \alpha} \ge 0$ and $c_{\alpha} > 0$. An important member of this class is the  "tree-reweighted" (TRW) approximation \cite{Wainwright-05upper} where $\bar c_\alpha$ is equal to a weighted combination of spanning trees of the original graph (all factors are pairwise and thus $\alpha$ represents an edge) which pass through $\alpha$. In Appendix~\ref{app:confree} we describe a number of concave settings of $\tilde H$ including TRW and other heuristic settings. The convex-free-energy variational program becomes:
\be
\min_{\bfb \in \LG} \sum_{i,x_i} b_i(x_i)\theta_i(x_i) + \sum_{\alpha,\bfx_\alpha} \theta_\alpha(\bfx_\alpha)b_\alpha(\bfx_\alpha) - \tilde H(\bfb).\label{eq:cfe}
\ee
The global minimizer $\bfb^*$ of the convex-free-energy program above is an approximation to the marginal probabilities due to (i) the term $\tilde H$ is an approximation to the entropy of the distribution and its quality depends on how the parameters $c_\alpha,c_i,c_{i\alpha}$ are set and on the structure of the factor graph, and (ii) due to the fact that the marginal consistency constraints $\LG$ approximate the probability simplex constraints, there is no guarantee that in general $\bfb^*$ form a distribution, i.e., the marginal estimations $b^*_\alpha(\bfx_\alpha)$ and $b^*_i(x_i)$ might not arise from any probability distribution over $x_1,...,x_n$.

The only guarantees we have is that if the factor graph has no cycles then the marginal probabilities are exact and if $\tilde H$ is strictly concave then it should be possible to generate a convergent message-passing algorithm (unlike BP algorithms which are not generally convergent).

We move next to the MAP assignment task where one seeks a  vector $\bfx^*$ which maximizes the product of potentials, or equivalently minimizes the energy

{\small $$\argmaxx{\bfx} \prod_i \phi_i(x_i) \prod_\alpha \psi_\alpha (\bfx_\alpha) = \argminn{\bfx} \sum_i \theta_i(x_i) + \sum_\alpha \theta_\alpha(x_\alpha).$$}
Described as a variational principle program, the MAP assignment problem is equivalent to the linear program whose variables corresponds to distribution $p(x_1,...,x_n)$ with exponential many elements:
$$ \min_{p(\bfx) \ge 0, \sum_{\bfx} p(\bfx)=1} \sum_{i,x_i} \theta_i(x_i) p(x_i) + \sum_{\alpha, \bfx_\alpha} \theta_\alpha(\bfx_\alpha) p(\bfx_\alpha).$$
The optimization of a linear function over the probability simplex yields an optimal solution $\bfp^*$ in an extreme point of the probability simplex, namely $\bfp^*$ is a zero-one distribution. In particular $p^*(\bfx^*)=1$ and for every $\bfx \ne \bfx^*$ holds $p^*(\bfx)=0$.

An approximation can be obtained by approximating the marginal probabilities $p(x_i)$ and $p(\bfx_\alpha)$ with beliefs $b_i(x_i)$ and $b_\alpha(\bfx_\alpha)$ which are not guaranteed to correspond to a true distribution over $x_1,...,x_n$.

%to an integer Linear Programming (LP) problem:
%$$ \min_{\bfb\in \LG} \sum_\alpha\sum_{\bfx_\alpha} \theta_\alpha(\bfx_\alpha)b_\alpha(\bfx_\alpha) \hspace{0.25cm} \mbox{s.t} \hspace{0.25cm} b_\alpha(x_\alpha) \in \{0,1\}.$$
%The integer problem is  equivalent to the original MAP problem, and is hence computationally intractable. An approximation can be obtained by relaxing the integer constraints whereby the beliefs take non-integer values. That is, we drop  the integral constraint $b_\alpha(x_\alpha) \in \{0,1\}$:

\be
\label{eq:LP}
\min_{\bfb_i, \bfb_\alpha \in \LG} \sum_{i,x_i} \theta_i(x_i) b_i(x_i) + \sum_{\alpha, \bfx_\alpha} \theta_\alpha(\bfx_\alpha)b_\alpha(\bfx_\alpha)
\ee
If the minimizer of the LP-relaxation problem comes out without ties, i.e., the marginal vectors $b_i(x_i)$ have a single maximal entry, then the MAP assignment readily emerges from the LP-relaxed solution. This LP-relaxed problem can be solved using off-the-shelf LP solvers but the key problem with standard LP solvers, however, is that they do not use the graph structure explicitly and thus are sub-optimal in terms of computational efficiency. An empirical study found the message-passing LP-solvers, e.g. max-TRBP, to be superior to the CPLEX solver, a commercial LP solver that implements different techniques for solving LP, such as primal and dual simplex solvers, network solvers, primal-dual barrier solver for sparse problem, and sifting techniques executing sequences of LP subproblems \cite{Yanover-jmlr06}. 

The relaxed LP problem of eqn.~\ref{eq:LP} has been widely studied in the literature in the context of message-passing algorithms. Special cases of these LP-relaxations were used for constraints satisfaction \cite{Schlesinger76}, \cite{Koster98}. The general form in eqn.~\ref{eq:LP} was studied using tree decompositions in \cite{Wainwright-05map},  \cite{Kolmogorov-pami06}, as well as dual decomposition \cite{Komodakis-eccv08}, \cite{Komodakis-pami10}, and dual block coordinate ascent \cite{Werner07}, \cite{Globerson-nips07}, \cite{Sontag-aistats09}. A general framework for these recent developments is described in  \cite{Meltzer-uai09}. Since the LP energy is not strictly convex, convergence to the global minimum is a challenge, since eqn. \ref{eq:LP} usually corresponds to a non-smooth dual. In this case a dual block coordinate ascent can lead to a corner in the dual objective, which is a non-optimal stationary point. 

An alternative class of methods are based on a (strictly) convex relaxation approach. There are two notable recent examples in this class: one using a proximal minimization technique where the convex term is a weighted KL-divergence measure between the sought-after belief vector and the one from the previous iteration \cite{Ravikumar-etal-icml08}. The proximal minimization approach involves a double-loop of message passing iterations and is guaranteed to converge to the global optimum of eqn.~\ref{eq:LP}. The second approach, the one we follow in this paper, is to make eqn.~\ref{eq:LP} the "zero temperature" of the perturbed problem:
\be
\min_{\bfb \in \LG} \sum_{i,x_i} \theta_i(x_i) b_i(x_i) + \sum_{\alpha,\bfx_\alpha} \theta_\alpha(\bfx_\alpha)b_\alpha(\bfx_\alpha) - \epsilon\tilde H(\bfb),\label{eq:LPe}
\ee
by taking $\epsilon\rightarrow 0$. This approach was used in decoding low-density parity-check codes \cite{Vontobel06}. It was also used for LP-relaxations, to derive a non-convergent max-product like algorithm \cite{Weiss-uai07}, and for applying an iterative proportional fitting type algorithm \cite{Johnson07}. 

This concludes the necessary background to inference within the framework of variational principle. The variational problem we will work on next is eqn.~\ref{eq:LPe}. We will derive a convergent message-passing algorithm called the norm-product. When the parameters of $\tilde H$ are set to the Bethe approximation the algorithm reduces to the sum-product (when $\epsilon=1$) or the max-product (when $\epsilon=0$). When $\tilde H$ is concave and $\epsilon=1$ the norm-product becomes a globally convergent message-passing algorithm, referred to as convex-sum-product,  for approximating marginal probabilities. When $\epsilon=0$ we obtain a convergent form of max-product we call convex-max-product and when $\epsilon\rightarrow 0$ we obtain an approximation (with proven bounds) to the LP-relaxation solution.

\section{The Norm-Product Belief Propagation Algorithm}
\label{sec:np}

We seek an algorithm for minimizing the inference variational eqn.~\ref{eq:LPe} with the following properties: (i) if the entropy approximation term $\tilde H$ is strictly concave, i.e., eqn.~\ref{eq:LPe} is a convex-free-energy, the algorithm will be convergent for all $\epsilon\ge 0$ and will converge to the global optimum when $\epsilon>0$, (ii) the algorithm will remain well defined when $\tilde H$ is non-convex (such as Bethe-free-energy and other fractional entropy approximations) and exhibit the property that fixed points of the algorithm coincide with stationary points of eqn.~\ref{eq:LPe}, and (iii) the algorithm uses the graph structure inherent sparseness, i.e., is defined by a message-passing architecture on the underlying factor-graph. In other words, like the BP-algorithms, our scheme should be sending messages between variable and factor nodes of the factor graph.

We will first take a detour and derive a general framework for minimizing problems of the type
\be
\min_{\bfb} f(\bfb) + \sum_{i=1}^n h_{ i} (\bfb)\label{eq:fh-general}
\ee
$f(\bfb)$ is a strictly convex, non-smooth, extended-valued function of the type $f(\bfb) = \hat f(\bfb) + \delta_{B}(\bfb)$ where $\hat f$ is essentially smooth and $\delta_{\cal B}$ is the indicator function of the affine set ${\cal B}=\{\bfb\ :\ A\bfb=\bfc\}$, namely, $\delta_{\cal B}(\bfb)=0$ if $\bfb\in {\cal B}$ and $\infty$ otherwise. The functions $h_i(\bfb)$ are convex extended-valued functions (see Appendix~\ref{app:bck} on mathematical background). In Appendix~\ref{app:dykstra} we derive the following "primal-dual" block ascent algorithm which is guaranteed to converge to the global minimizer of eqn.~\ref{eq:fh-general}: \vspace{0.2cm}

\begin{algorithm}[Primal-Dual Ascent]
\label{alg:pd-general}
Let $f(\bfb) = \hat f(\bfb) + \delta_{\cal B}(\bfb)$ where $\hat f(\bfb)$ is strictly convex, essentially smooth extended-valued function, and let $h_i(\bfb)$ be convex extended-valued functions.
Initialize $\bflambda_1=\bfzero,...,\bflambda_n=\bfzero$.
\begin{enumerate}
\item Repeat until convergence:
\item For $i=1,...n$:
  \begin{enumerate}
  \item $\bfmu_i =   \sum_{j \ne i} \bflambda_j$
  \item $\bfb^* =  \argminn{ \bfb \in dom(h_i)\cap dom(f)}\left\{f(\bfb) + h_i(\bfb) + \bfb^\top \bfmu_i \right\}$
  \item $\bflambda_i =  -\bfmu_i - \nabla \hat f(\bfb^*) + A^\top \bfsigma$ where $\bfsigma$ is arbitrary.
  \end{enumerate}
\end{enumerate}
Output $\bfb^*$.
\end{algorithm} \vspace{0.2cm}

\begin{figure*}
\fbox{
\begin{minipage}[c]{17.5cm}
\begin{algorithm}[Norm-Product Belief Propagation]
\label{alg:np}
We are given nonnegative local evidence $\phi_i(x_i)$, and nonnegative arrays $\psi_{\alpha}(\bfx_{\alpha})$, where $\alpha \subset \{1,...,n\}$. Let $ \hat c_{i\alpha} = c_\alpha + c_{i\alpha}$ and $\hat c_i = c_i + \sum_{\alpha\in N(i)} c_\alpha$.
\begin{enumerate}
\item Set $n_{i  \rightarrow \alpha}(\bfx_\alpha) =1$ for all $i=1,...,n$, $\alpha\in N(i)$ and $\bfx_\alpha$.
\item For $t=1,2,...$
\begin{enumerate}
\item For $i=1,...n$ do:
\beas
\forall x_i \; \forall \alpha \in N(i) \;\;\;\; m_{\alpha\rightarrow i}(x_i)&=&\left( \sum_{\bfx_\alpha\setminus x_i} \left( \psi_\alpha(\bfx_\alpha)\prod_{j\in N(\alpha)\setminus i}n_{j\rightarrow\alpha}(\bfx_\alpha)\right)^{1/(\epsilon\hat c_{i\alpha})} \right)^{\epsilon\hat c_{i\alpha}}\\
%\forall x_i \; \forall \alpha \in N(i) \;\;\;\; m_{\alpha\rightarrow i}(x_i)&=&  \underset{\bfx_\alpha\setminus x_i}{\Big\|} \psi_\alpha(\bfx_\alpha)\prod_{j\in N(\alpha)\setminus i}n_{j\rightarrow\alpha}(\bfx_\alpha) \Big\|_{1/(\epsilon\hat c_{i\alpha})}\\
\forall \alpha \in N(i) \; \forall \bfx_\alpha \;\;\;\;  n_{i\rightarrow\alpha}(\bfx_\alpha)&\propto&  \left(\frac{\displaystyle \phi^{1/\hat c_i}_i(x_i) \prod_{\beta\in N(i)}m_{\beta\rightarrow i}^{1/\hat c_i}(x_i)}{m_{\alpha\rightarrow i}^{1/\hat c_{i\alpha}}(x_i)}\right)^{c_\alpha}\left(\psi_\alpha(\bfx_\alpha)\prod_{j\in N(\alpha)\setminus i}n_{j\rightarrow\alpha}(\bfx_\alpha)\right)^{-c_{i\alpha}/\hat c_{i\alpha}}
\eeas
\end{enumerate}
\end{enumerate}
\end{algorithm}
\end{minipage}
}
\caption{The norm-product belief propagation algorithm, where the messages $m_{\alpha \rightarrow i}(x_i)$ are computed with respect to the $L_{1/\epsilon \hat c_{i \alpha}}$ norm. For $c_\alpha=1, c_i=1-d_i, c_{i \alpha}=0$ it reduces to the belief propagation algorithms, sum-product when $\epsilon=1$ and max-product when $\epsilon=0$. Whenever $c_\alpha$ is the weighted number of spanning trees through edge $\alpha$, and $c_i = 1-\sum_{\alpha \in N(i)} c_\alpha$ and $c_{i \alpha}=0$ it reduces to the tree-reweighted belief propagation algorithms (sum-TRBP and max-TRBP). Whenever $c_\alpha>0, c_i, c_{i \alpha} \ge 0$ the norm-product is guaranteed to converge, and if also $\epsilon > 0$ it converges to the global optimum of the program in eqn. \ref{eq:LPe}.}
\label{fig:norm-prod-alg}
\end{figure*}

The vectors $\bflambda_i$ and $\bfmu_i$ are messages passed along edges of a bipartite graph with $n$ (function) nodes corresponding to the $n$ functions $h_i(\bfb)$ and $m$ (variable) nodes corresponding to the dimension of $\bfb$. Function node $i$ sends the $m$ coordinates of vector $\bflambda_i$ to the $m$ variable nodes. Variable node $j$ sends the $j$'th coordinate of vectors $\bfmu_1,...,\bfmu_n$ to the $n$ functions nodes. %(see Fig.~\ref{fig:messages}). 
The algorithm iteratively optimizes with respect to the indexes $i \in \{1,...,n\}$, stopping when it does not change the beliefs $\bfb^*$, thus the network proceeds in an almost cyclic update policy. The algorithm fits well with a graphical model architecture in the sense that if $h_i(\bfb)$ depends only on a small subset $N(i)$ of coordinates from $\bfb$, then $\bflambda_{i \beta}=0$ for every $\beta \not \in N(i)$ (and therefore need not be updated): \vspace{0.2cm}
\begin{claim}
\label{claim:l-sparse}
Assume variables $\bfb$ are indexed by $\{1,...,m\}$ and $h_i(\bfb)$ depends only on small subset of variables indexed by $N(i) \subset \{1,...,m\}$ and let  $\bflambda_i=\{\bflambda_{i,\alpha}\}$. Then, $\bflambda_{i,\beta}=\bfzero$ for all $\beta \not\in N(i)$.
\end{claim} \vspace{0.1cm} 
The claim and its proof can be found in Appendix~\ref{app:dykstra}.
For those familiar with successive projection schemes, in the particular case when $f(\bfb)=\hat f(\bfb)$, i.e., is strictly convex and essentially smooth, and $h_i(\bfb)=\delta_{C_i}(\bfb)$ (the indicator function of convex set $C_i$), the update step (b) for Algorithm \ref{alg:pd-general} is a "Bregman" projection \cite{bregman99} of the vector $\bfmu_i$ onto the convex set $C_i$. In that case, following some algebraic manipulations (such as eliminating $\bfmu_i$ among other manipulations) the scheme (with $A=0$) reduces to the well known Dykstra \cite{Dykstra83} (also goes under different names such as Hildreth, Bregman, Csiszar, Han) successive projection algorithm which has its origins in the work of Von-Neumann \cite{Von-Neumann50}. Further historical details can be found in Appendix~\ref{app:dykstra}.

Another useful property of the algorithm that it is well defined for non-convex primal energies. Specifically, we can establish the following result: \vspace{0.1cm} 
\begin{claim}
\label{claim:non-convex}
Consider Algorithm \ref{alg:pd-general} for Legendre-type function $f(\bfb)$ and non-convex continuously differentiable functions $h_i(\bfb)$ restricted to the affine domain $\{\bfb\ :\ A_i \bfb = \bfc_i\}$, and assume $\bfb^*$ in step (c) is in the interior of $dom(f)$. Then, fixed-points of the algorithm coincide with stationary points of the non-convex program $f(\bfb) + \sum_i h_i(\bfb)$.
\end{claim} \vspace{0.2cm} 
The proof can be found in Appendix~\ref{app:dykstra-nc}. The result states that when $h_i$ are non-convex but defined over an affine domain the algorithm is no longer convergent, but if it does converge it will do so to a stationary point of the optimization problem. This property of the algorithm extends the result of \cite{Yedidia05} about the behavior of the sum-product algorithm: if it converges, then it converges to a stationary point of the Bethe free-energy.

The inference variational problem presented in eqn.~\ref{eq:LPe} is embedded into the general template of eqn.~\ref{eq:fh-general} as follows: 
\be
\min_{\bfb} f_{\epsilon}(\bfb) + \sum_{i=1}^n h_{\epsilon, i} (\bfb)\label{eq:fh}
\ee
where
$f_{\epsilon}(\bfb)=\hat f_{\epsilon}(\bfb) + \delta_S(\bfb)$ with $S$ being the set of $\{\bfb=\{\bfb_\alpha\}\ :\ b_\alpha(\bfx_\alpha)\in {\cal P}\}$ where $\cal P$ is the probability simplex  (arrays that are non-negative and sum to one), and $\hat f_{\epsilon}$ is defined below:
%$$S=\{\bfb=\{\bfb_\alpha\}_\alpha\ :\ \bfb\ge 0, \sum_{\bfx_\alpha}b_\alpha(\bfx_\alpha)=1\},$$
%and 
\be  \hat f_{\epsilon}(\bfb_\alpha)=\sum_{\alpha,\bfx_\alpha} \theta_\alpha(\bfx_\alpha)b_\alpha(\bfx_\alpha) - \sum_\alpha \epsilon c_{\alpha} H(\bfb_\alpha)\label{eq:f}
\ee
Note that $dom(f_{\epsilon})$ include all $\bfb\in S$, i.e., $f_{\epsilon}(\bfb)=\infty$ for $\bfb\not\in S$. The functions $h_{\epsilon, i}$ are defined below:
{\small \be
\label{eq:hi}
h_{\epsilon, i}(\bfb)= \sum_{x_i} \theta_i(x_i) b_i(x_i) - \epsilon  c_i H(\bfb_i) - \hspace{-0.2cm} \sum_{\alpha\in N(i)} \hspace{-0.3cm} \epsilon c_{i \alpha} (H(\bfb_\alpha) - H(\bfb_i)),
\ee}

where $dom(h_{\epsilon, i})$ is the affine set consisting of $\bfb_\alpha$ for every $\alpha\in N(i)$, which live in the probability simplex, i.e.  $dom(h_{\epsilon, i})\subset dom(f_{\epsilon})$, and satisfy the marginal consistency constraints $ \sum_{\bfx_\alpha\setminus  x_i}b_\alpha(\bfx_\alpha) = b_i(x_i)$. Note that $\bfb_i$ are not explicitly included in  $\bfb = \{\bfb_\alpha\}$, but they are described by the values of which all $\bfb_\alpha$ in the domain of $h_{\epsilon,i}$ agree upon. 

Given the sparse structure of $h_{\epsilon,i}$  then, following Claim~\ref{claim:l-sparse}, we present the entries of $\bflambda_i$ according to the factor-graph structure by setting  $\bflambda_{i}=\{\lambda_{i,\alpha}(\bfx_\alpha)\}$ (and likewise $\bfmu_i$). Step (b) of Algorithm~\ref{alg:pd-general} is reduced to finding $\bfb^*_\alpha$ for all $\alpha\in N(i)$ and step (c) updates $\bflambda_{i,\alpha}$ by the rule:
$$\lambda_{i,\alpha}(\bfx_\alpha) =  -\mu_{i,\alpha}(\bfx_\alpha) - \nabla \hat f_\epsilon(b^*_\alpha(\bfx_\alpha)) + \sigma_\alpha\bfone,$$
for an arbitrary $\sigma_\alpha$. 
If instead of updating $\bflambda_{i,\alpha}$ we would update $n_{i\rightarrow\alpha}(\bfx_\alpha)=\exp(-\lambda_{i,\alpha}(\bfx_\alpha))$ the additive degree of freedom inherent in the choice of $\bfsigma$ turns into a scaling choice of $n_{i\rightarrow\alpha}$.

The derivation process required for embedding the definitions above into the primal dual Algorithm~\ref{alg:pd-general} is described in detail in Appendix~\ref{app:np}. The resulting algorithm, we call {\it norm-product\/}, is presented in Fig.~\ref{fig:norm-prod-alg}.

Just as in the BP algorithms, the {\it message\/} $m_{\alpha\rightarrow i}(x_i)$ from the factor node $\alpha$ to the variable node $i$ is a vector over all possible states of $ x_i$. The message $n_{i\rightarrow\alpha}(\bfx_\alpha)$ from the variable node $i$ to the factor node $\alpha$ is an {\it array\/} over all possible states of $\bfx_\alpha$. The beliefs $b_i(x_i)$, which are the approximations to the marginal probability $p(x_i)$ when $\epsilon=1$, can be computed from the messages $m_{\alpha\rightarrow i}$:
\be
b_i(x_i)\propto \left( \phi_i(x_i) \prod_{\alpha\in N(i)}m_{\alpha\rightarrow i}(x_i)\right)^{1/\epsilon\hat c_i},\label{eq:bi2}
\ee
where $\hat c_i$ is defined in Fig.~\ref{fig:norm-prod-alg}. The joint beliefs $b_\alpha(\bfx_\alpha)$ can be computed from the messages $n_{i\rightarrow\alpha}$:
\be
b_\alpha(\bfx_\alpha)\propto  \left(\psi_\alpha(\bfx_\alpha)\prod_{i\in N(\alpha)}n_{i\rightarrow\alpha}(\bfx_\alpha) \right)^{1/\epsilon c_\alpha}.\label{eq:balpha2}
\ee

The norm-product algorithm includes the BP algorithms (sum-product and max-product), as well as sum-TRBP \cite{Wainwright-05upper}, max-TRBP \cite{Wainwright-05map}, and NMPLP \cite{Globerson-nips07} as particular cases. These algorithms relate to the simpler form of the norm-product algorithm, when $c_{i \alpha}=0$. In this setting the messages $n_{i \rightarrow \alpha}(\bfx_\alpha)$ depend solely on the local potentials $\phi_i(x_i)$ and the messages $m_{\beta \rightarrow i}(x_i)$. Therefore the messages $n_{i \rightarrow \alpha}(\bfx_\alpha)$ can be written in the compact form $n_{i \rightarrow \alpha}(x_i)$, replacing $\bfx_\alpha$ with $x_i$. In this case the norm-product algorithm in Fig. \ref{fig:norm-prod-alg} takes the form:

{\small 
\beas
%m_{\alpha\rightarrow i}(x_i)&=& \underset{\bfx_\alpha\setminus x_i}{\Big\|} \psi_\alpha(\bfx_\alpha)\prod_{j\in N(\alpha)\setminus i}n_{j\rightarrow\alpha}(x_j) \Big\|_{1/\epsilon c_\alpha} \\
m_{\alpha\rightarrow i}(x_i) \hspace{-0.15cm}  &=& \hspace{-0.15cm}  \left( \sum_{ \bfx_\alpha\setminus x_i} \left( \psi_\alpha(\bfx_\alpha)\prod_{j\in N(\alpha)\setminus i}n_{j\rightarrow\alpha}(\bfx_\alpha)\right)^{1/\epsilon c_\alpha} \right)^{\epsilon c_\alpha} \\
n_{i\rightarrow\alpha}(x_i)&\propto&  \frac{\left(\phi_i(x_i) \displaystyle \hspace{-0.15cm} \prod_{\beta\in N(i)\setminus \alpha} \hspace{-0.3cm} m_{\beta\rightarrow i}(x_i) \right)^{c_\alpha/\hat c_i}}{m_{\alpha\rightarrow i}(x_i)}
\eeas
}
When using the norm-product with the Bethe entropy approximation $c_{i \alpha}=0, c_\alpha=1, c_i=1-d_i$ there holds $\hat c_i =1$ and the algorithm reduces to 

{\small 
\beas
%m_{\alpha\rightarrow i}(x_i)&=& \underset{\bfx_\alpha\setminus x_i}{\Big\|} \psi_\alpha(\bfx_\alpha)\prod_{j\in N(\alpha)\setminus i}n_{j\rightarrow\alpha}(x_j) \Big\|_{1/\epsilon} \\
m_{\alpha\rightarrow i}(x_i)&=&  \left( \sum_{ \bfx_\alpha\setminus x_i} \left( \psi_\alpha(\bfx_\alpha)\prod_{j\in N(\alpha)\setminus i}n_{j\rightarrow\alpha}(\bfx_\alpha)\right)^{1/\epsilon} \right)^{\epsilon} \\
n_{i\rightarrow\alpha}(x_i)&\propto&  \phi_i(x_i) \prod_{\beta\in N(i)\setminus \alpha}m_{\beta\rightarrow i}(x_i)
\eeas
}
which is the sum-product algorithm for $\epsilon=1$ and the max-product algorithm for $\epsilon=0$. 
%
%The sum-product algorithm arises from the setting $\epsilon=1$ and the setting of $\tilde H$ as the Bethe entropy approximation: $c_\alpha=1, c_i=1-d_i$ and $c_{i\alpha}=0$ from which we obtain $\hat c_{i\alpha}=\hat c_i = 1$. Substituting back into the message definitions we obtain:
%\beas
%m_{\alpha\rightarrow i}(x_i)&=&\sum_{ \bfx_\alpha\setminus x_i}\psi_\alpha(\bfx_\alpha)\prod_{j\in N(\alpha)\setminus i}n_{j\rightarrow\alpha}(x_j)\\
%\eeas
%which are exactly the sum-product messages. Note that the message $n_{i\rightarrow\alpha}(\bfx_\alpha)$ remains only with $ x_i$ components (because $c_{i\alpha}=0$) and therefore becomes $n_{i\rightarrow\alpha}(x_i)$. The joint belief $b_\alpha(\bfx_\alpha)$ reduces to:
%$$b_\alpha(\bfx_\alpha)\propto  \psi_\alpha(\bfx_\alpha)\prod_{j\in N(\alpha)}n_{j\rightarrow\alpha}(x_j).$$
%The max-product algorithm arises from the setting $\epsilon=0$ (and $\tilde H$ the Bethe entropy approximation). The $L_{1/\epsilon}$ norm in the definition of the message $m_{\alpha\rightarrow i}(x_i)$ becomes an $L_\infty$ norm:
%\be
%m_{\alpha\rightarrow i}(x_i)=\max_{ \bfx_\alpha\setminus x_i}\left\{\psi_\alpha(\bfx_\alpha)\prod_{j\in N(\alpha)\setminus i}n_{j\rightarrow\alpha}(x_j)\right\}\label{eq:mp-msg}
%\ee
%The belief $b_i(x_i)$ to the power of $\epsilon$ is computed from:
%$$b^\epsilon_i(x_i)\propto \phi_i(x_i) \prod_{\alpha\in N(i)}m_{\alpha\rightarrow i}(x_i),$$
%which is a binary vector (since $\epsilon=0$) representing the MAP solution (if there are no ties). 

\begin{figure*}
\fbox{
\begin{minipage}[c]{17.5cm}
\begin{algorithm}[Sum-Product Belief Propagation type]
\label{alg:csp}
We are given nonnegative local evidence $\phi_i(x_i)$, and nonnegative arrays $\psi_{\alpha}(\bfx_{\alpha})$, where $\alpha \subset \{1,...,n\}$. Let $ \hat c_{i\alpha} = c_\alpha + c_{i\alpha}$ and $\hat c_i = c_i + \sum_{\alpha\in N(i)} c_\alpha$.
\begin{enumerate}
\item Set $n_{i  \rightarrow \alpha}(\bfx_\alpha) =1$ for all $i=1,...,n$, $\alpha\in N(i)$ and $\bfx_\alpha$.
\item For $t=1,2,...$
\begin{enumerate}
\item For $i=1,...n$ do:
\beas
\forall x_i \; \forall \alpha \in N(i) \;\;\;\; m_{\alpha\rightarrow i}(x_i)&=& \left( \sum_{ \bfx_\alpha\setminus x_i} \left( \psi_\alpha(\bfx_\alpha)\prod_{j\in N(\alpha)\setminus i}n_{j\rightarrow\alpha}(\bfx_\alpha)\right)^{1/\hat c_{i\alpha}} \right)^{\hat c_{i\alpha}} \\
%\forall x_i \; \forall \alpha \in N(i) \;\;\;\; m_{\alpha\rightarrow i}(x_i)&=&  \underset{\bfx_\alpha\setminus x_i}{\Big\|} \psi_\alpha(\bfx_\alpha)\prod_{j\in N(\alpha)\setminus i}n_{j\rightarrow\alpha}(\bfx_\alpha) \Big\|_{1/(\hat c_{i\alpha})}\\
\forall \alpha \in N(i) \; \forall \bfx_\alpha \;\;\;\; n_{i\rightarrow\alpha}(\bfx_\alpha)&\propto&  \left(\frac{\displaystyle \phi_i^{1/\hat c_i}(x_i) \prod_{\beta\in N(i)}m_{\beta\rightarrow i}^{1/\hat c_i}(x_i)}{m_{\alpha\rightarrow i}^{1/\hat c_{i\alpha}}(x_i)}\right)^{c_\alpha}\left(\psi_\alpha(\bfx_\alpha)\prod_{j\in N(\alpha)\setminus i}n_{j\rightarrow\alpha}(\bfx_\alpha)\right)^{-c_{i\alpha}/\hat c_{i\alpha}}
\eeas
\end{enumerate}
\end{enumerate}
\end{algorithm}
\end{minipage}
}
\caption{Sum-product belief propagation type algorithm, attained from the norm-product belief propagation when $\epsilon=1$, where the messages $m_{\alpha \rightarrow i}(x_i)$ are computed with the $L_{1/ \hat c_{i \alpha}}$ norm. For $c_\alpha=1, c_i=1-d_i, c_{i \alpha}=0$ it reduces to the sum-product belief propagation algorithms, and  whenever $c_\alpha$ is the weighted number of spanning trees through edge $\alpha$, and $c_i = 1-\sum_{\alpha \in N(i)} c_\alpha$ and $c_{i \alpha}=0$ it reduces to sum-TRBP. If $c_\alpha>0, c_i, c_{i \alpha} \ge 0$ it reduces to the convex-sum-product algorithm, which is guaranteed to reach the global optimum of the convex free energy.} 
\label{fig:convex-sum-prod}
\end{figure*}

When the factors corresponds to pairwise interactions $\alpha = (i,j)$ the messages of norm-product algorithm $m_{\alpha \rightarrow i}$ and $n_{i \rightarrow \alpha}$ can be written by the shorthand notation $m_{j \rightarrow i}$ and $n_{i \rightarrow j}$. The messages $m_{j \rightarrow i}$ of the norm-product algorithm in Fig.~\ref{fig:norm-prod-alg} depends on a single message $n_{j \rightarrow i}$ and whenever $c_{i \alpha}=0$ the message $n_{j \rightarrow i}$ depends only on the messages $m_{k \rightarrow j}$ for every $\{k,j\} \in N(j)$, which we abbreviate by $k \in N(j)$. Substituting the value of $n_{j \rightarrow i}$ into $m_{j \rightarrow i}$ we obtain the pairwise norm-product, whose update rule consists only of the messages $m_{k \rightarrow j}$. When $\epsilon=1$ the pairwise norm-product algorithm with $c_{i \alpha}=0$ takes the form  $$m^{1/c_{ij}}_{j \rightarrow i}(x_i) \propto \sum_{x_j} \psi^{1/c_{ij}}_{ij}(x_i,x_j)  \frac{\phi_j^{1/\hat c_j}(x_j) \prod_{k \in N(j)} m_{k \rightarrow j}^{1/\hat c_j}(x_j)}{m^{1/c_{ij}}_{i \rightarrow j}(x_j)}.$$ 
The sum-TRBP \cite{Wainwright-05upper} is a special case. The sum-TRBP sets $c_{ij}$ as the relative number of spanning trees of the graph which include the edge $(i,j)$, and sets $c_i=1-\sum_{j \in N(i)} c_{ij}$. As a result $\hat c_i = 1$ and 
%the update message of the pairwise norm-product becomes:
%$$m^{1/c_{ij}}_{j \rightarrow i}(x_i) \propto \sum_{x_j} \psi^{1/c_{ij}}_{ij}(x_i,x_j)  \frac{\phi_j(x_j) \prod_{k \in N(j)} m_{k \rightarrow j}(x_j)}{m^{1/c_{ij}}_{i \rightarrow j}(x_j)}.$$
by substitution  $M_{ij}(x_i) \stackrel{def}{=} m^{1/c_{ij}}_{j \rightarrow i}(x_i)$ we obtain the sum-TRBP update rule as originally introduced in (\cite{Wainwright-05upper}, eqn. 39):
$$M_{ij}(x_i) \propto \sum_{x_j} \psi^{1/c_{ij}}_{ij}(x_i,x_j)  \frac{\phi_j(x_j) \prod_{k \in N(j)} M^{c_{jk}}_{jk}(x_j)}{M_{ji}(x_j)}.$$

When $\epsilon=0$ the pairwise norm-product algorithm with $c_{i \alpha}=0$ takes the form  $$m_{j \rightarrow i}(x_i) \propto \max_{x_j} \psi_{ij}(x_i,x_j)  \frac{\phi_j^{c_{ij}/\hat c_j}(x_j) \prod_{k \in N(j)} m_{k \rightarrow j}^{c_{ij}/\hat c_j}(x_j)}{m_{i \rightarrow j}(x_j)}.$$ 
The max-TRBP \cite{Wainwright-05map}  and NMPLP \cite{Globerson-nips07} are special cases, derived as follows: With max-TRBP, we have $c_{ij}$ and $c_i$ defined by the tree-reweighted setting which results in $\hat c_i=1$, and the Max-TRBP (\cite{Wainwright-05map}, eqn. 50) follows from the substitution $M_{ij}(x_i) \stackrel{def}{=} m_{j \rightarrow i}^{1/c_{ij}}(x_i)$. 
The NMPLP  is another recent max-product-like algorithm where  messages $\gamma_{ji}(x_i)$ are defined as follows:
\be
\gamma_{ji}(x_i)=\max_{\bfx_j}\left\{
\theta_{ij}(x_i,x_j)-\gamma_{ij}(x_j)+w_j \sum_{k\in N(j)}\gamma_{kj}(x_j)\right\}\nonumber
\ee
where $w_j=2/(d_j+1)$. The pairwise norm-product message $m_{j \rightarrow i}(x_i)$ with the setting 
$c_j=(1 - d_j)/2$ and $c_{ij}=1$ for every $(i,j)$ gives rise to $c_{ij}/\hat c_j = 2/(d_j+1)$. Thus with the substitution $\gamma_{ji}(x_i) \stackrel{def}{=} \ln m_{j \rightarrow i}(x_i)$ and unit local potentials $(\phi_i(x_i)=1)$ we obtain the NMPLP message above.

%It is worthwhile noting that since $c_i$ is negative this setting of parameters is not written as a convex-free-energy thus in general, like with BP algorithms, for the case of $\epsilon=1$ we can rely on Claim~\ref{claim:non-convex} to assert the property that (i) the sum-TRBP is not guaranteed to converge, but (ii) if it does converge it will do so to a stationary point of the TRBP-free-energy. As for the max-TRBP, we can guarantee that upon convergence the marginal consistency constraints are satisfied for {\it some\/} spanning tree (as already shown by \cite{Wainwright-05map}) but we cannot guarantee convergence. The situation with convex-free-energies (and TRW-free-energy can be recast as a convex energy as described in Appendix~\ref{app:confree}) is much more encouraging: we will show in Section~\ref{sec:cmp} that the setting of convex-free-energy with $\epsilon=0$ is convergence guaranteed. On the other hand, using different proof, the NMPLP is guaranteed to converge \cite{Globerson-nips07}. 

%

The result of having the BP, TRBP and NMPLP algorithms arise as special cases of the norm-product algorithm underscores the generality of our derivation. However, the more interesting potential in the norm-product algorithm is the emergence of {\it new\/} message-passing schemes which are guaranteed to converge (unlike the BP and TRBP algorithms) corresponding to the setting of $\tilde H$ as a concave function ($c_\alpha>0, c_i,c_{i\alpha}\ge 0$). Three classes of algorithms emerge: \begin{itemize}
\item The {\it convex-sum-product\/} corresponding to the setting $\epsilon=1$ in the norm-product algorithm. The convex-sum-product is guaranteed to converge to the global optimum of the primal function eqn.~\ref{eq:LPe}. This includes the tree-reweighted free-energy in particular and other settings of convex-free-energy which are detailed in Appendix~\ref{app:confree}.
\item The {\it approximate LP-relaxation\/} corresponding to the setting $\epsilon\rightarrow 0$ (but $\epsilon >0$)  in the norm-product algorithm. It provides an approximate solution to the LP-relaxation whose distance from the true solution is governed by an upper-bound we derive. The approximate LP-relaxation is guaranteed to converge to the global optimum of the primal function eqn.~\ref{eq:LPe}.
\item The {\it convex-max-product\/} corresponding to the setting $\epsilon=0$ in the norm-product algorithm. Unlike the max-product, the convex-max-product is convergence guaranteed. However, there is no guarantee that the recovered solution corresponds to the desired LP-relaxation solution.
The advantage of convex-max-product is efficiency (introduced by $L_\infty$ instead of $L_{1/\epsilon}$) and very good empirical performance. In fact, the convex-max-product is a convergent form of max-product.
\end{itemize}
These message-passing algorithms, which are collectively referred to as {\it convex-BP\/} algorithms, are discussed in the next section.

\section{The Convex Belief Propagation Algorithms}

Eqn.~\ref{eq:LPe} represents the free-energy approximation when $\epsilon=1$, the LP relaxation when $\epsilon=0$, and a perturbation of the LP-relaxation for MAP estimation when $\epsilon\rightarrow 0$. When the entropy approximation term $\tilde H$ is the Bethe approximation (setting $c_\alpha=1,c_i=1-d_i,c_{i\alpha}=0$ in eqn.~\ref{eq:H}) the sum-product ($\epsilon=1$) and max-product ($\epsilon=0$) arise as special cases of the norm-product algorithm. Since in both cases the free-energy approximation is non-convex (for factor graphs with cycles) the convergence guarantees of those algorithms are weak. For the sum-product we have the guarantee that {\it if\/} the algorithm convergence then it will reach a stationary point of the free-energy approximation (see Claim~\ref{claim:non-convex} and \cite{Yedidia05}). With the max-product we have weaker guarantees (Claim~\ref{claim:non-convex} does not apply because $f_{\epsilon}$ is not strictly convex when $\epsilon=0$) where specifically, even if the algorithm does converge the marginal consistency constraints might not be satisfied.

We focus now on the family of convex-free-energies which arise with the setting $c_\alpha>0, c_i,c_{i\alpha}\ge 0$.  The convex-sum-product arises from the setting $\epsilon=1$ is described next.

\subsection{Convex-sum-product Algorithm}

As a free-energy approximation ($\epsilon=1$), eqn.~\ref{eq:LPe} is strictly convex and, in turn, the norm-product algorithm is guaranteed to converge to the global optimum. We refer to the specialization of the norm-product algorithm with $c_\alpha>0, c_i,c_{i\alpha}\ge 0$ and $\epsilon=1$ as {\it convex-sum-product\/} summarized in Fig. \ref{fig:convex-sum-prod}. 

The beliefs $b_i(x_i)$, which are the approximations to the marginal probability $p(x_i)$, and the joint beliefs $b_\alpha(\bfx_\alpha)$, which are the approximation to the marginal probability $p(\bfx_\alpha)$, are computed from:
\beas
b_i(x_i)&\propto& \left(\phi_i(x_i) \prod_{\alpha\in N(i)}m_{\alpha\rightarrow i}(x_i)\right)^{1/\hat c_i},\\
b_\alpha(\bfx_\alpha)&\propto&  \left(\psi_\alpha(\bfx_\alpha)\prod_{j\in N(\alpha)}n_{j\rightarrow\alpha}(\bfx_\alpha)\right)^{1/c_\alpha}.
\eeas
Note that the algorithm has a much simpler form if $c_{i\alpha}=0$. The message $n_{i\rightarrow\alpha}(\bfx_\alpha)$ depends only on $ x_i$ and becomes:
\be
n_{i\rightarrow\alpha}(x_i)\propto  \frac{\displaystyle \left( \phi_i(x_i) \prod_{\beta\in N(i)}m_{\beta\rightarrow i}(x_i) \right)^{c_\alpha/\hat c_i} }{m_{\alpha\rightarrow i}(x_i)}. \label{eq:nia-short}
\ee
The convex-sum-product is globally convergent for any concave setting of the entropy approximation $\tilde H$, i.e., when $c_\alpha>0, c_i,c_{i\alpha}\ge 0$. In particular, when the underlying factor-graph arises from a graph, i.e., the local interaction forms pairwise relations only, there is a setting that corresponds to TRW free-energy as described in Appendix~\ref{app:confree}. We also describe there additional parameter settings corresponding to other heuristic convex approximations of the entropy term $\tilde H$.

We describe next the use of the norm-product algorithm as an approximation to the LP-relaxation for the MAP problem by taking $\epsilon\rightarrow 0$.

\begin{figure*}
\fbox{
\begin{minipage}[c]{17.5cm}
\begin{algorithm}[Max-Product Belief Propagation type]
\label{alg:cmp}
We are given nonnegative local evidence $\phi_i(x_i)$, and nonnegative arrays $\psi_{\alpha}(\bfx_{\alpha})$, where $\alpha \subset \{1,...,n\}$. Let $ \hat c_{i\alpha} = c_\alpha + c_{i\alpha}$ and $\hat c_i = c_i + \sum_{\alpha\in N(i)} c_\alpha$.
\begin{enumerate}
\item Set $n_{i  \rightarrow \alpha}(\bfx_\alpha) =1$ for all $i=1,...,n$, $\alpha\in N(i)$ and $\bfx_\alpha$.
\item For $t=1,2,...$
\begin{enumerate}
\item For $i=1,...n$ do:
\beas
\forall x_i \; \forall \alpha \in N(i) \;\;\;\; m_{\alpha\rightarrow i}(x_i)&=&\max_{ \bfx_\alpha\setminus x_i}\left\{\psi_\alpha(\bfx_\alpha)\prod_{j\in N(\alpha)\setminus i}n_{j\rightarrow\alpha}(\bfx_\alpha)\right\}\\
\forall \alpha \in N(i) \; \forall \bfx_\alpha \;\;\;\; n_{i\rightarrow\alpha}(\bfx_\alpha)&\propto&  \left(\frac{\displaystyle \phi_i^{1/\hat c_i}(x_i) \prod_{\beta\in N(i)}m_{\beta\rightarrow i}^{1/\hat c_i}(x_i)}{m_{\alpha\rightarrow i}^{1/\hat c_{i\alpha}}(x_i)}\right)^{c_\alpha}\left(\psi_\alpha(\bfx_\alpha)\prod_{j\in N(\alpha)\setminus i}n_{j\rightarrow\alpha}(\bfx_\alpha)\right)^{-c_{i\alpha}/\hat c_{i\alpha}}
\eeas
\end{enumerate}
%\item Output: $\hatb_i(x_i)\propto \prod_{\alpha\in N(i)}m_{\alpha\rightarrow i}(x_i)^{1/\hat c_i}$, $\ \ i=1,...,n$.
\end{enumerate}
\end{algorithm}
\end{minipage}
}
\caption{Max-product belief propagation type algorithm, attained from the norm-product belief propagation when $\epsilon=0$, where the messages $m_{\alpha \rightarrow i}(x_i)$ are computed with the $L_\infty$ norm. For $c_\alpha=1, c_i=1-d_i, c_{i \alpha}=0$ it reduces to the max-product belief propagation algorithms. Whenever $c_\alpha$ is the weighted number of spanning trees through edge $\alpha$, and $c_i = 1-\sum_{\alpha \in N(i)} c_\alpha$ and $c_{i \alpha}=0$ it reduces to max-TRBP. For $c_\alpha=1, c_i = (1-d_i)/2, c_{i \alpha}=0$ it reduces to the NMPLP algorithm. If $c_\alpha>0, c_i, c_{i \alpha} \ge 0$ it reduces to the convex-max-product algorithm, which is a convergent max-product type algorithm for LP-relaxations.}
\label{fig:convex-max-prod}
\end{figure*}

\subsection{LP-relaxation Bounds}
\label{sec:bounds}

For $\epsilon>0$, let the global optimum of eqn.~\ref{eq:LPe} (with concave $\tilde H$) denoted by $\bfb_\epsilon$ and let the solution of the LP relaxation eqn.~\ref{eq:LP} denoted by $\bfb^*$. 
Let $\bftheta$ stand for the concatenated functions $\theta_i(x_i)$ and $\theta_\alpha(\bfx_\alpha)$, i.e., $\bftheta^\top\bfb = \sum_{i, x_i} \theta_i(x_i)b_i(x_i) + \sum_{\alpha, \bfx_\alpha} \theta_\alpha(\bfx_\alpha)b_\alpha(\bfx_\alpha)$. We wish to upper-bound the difference $\bftheta^\top \bfb_\epsilon - \bftheta^\top \bfb^*\le \delta$ where $\delta$ is a function of $\epsilon, c_\alpha, c_i$ and $c_{i\alpha}$, described below: \vspace{0.1cm}
\begin{proposition}
\label{prop:bound}
Let  $c_\alpha>0, c_i,c_{i\alpha}\ge 0$ describe a convex-free-energy eqn.~\ref{eq:LPe}. Let $n_i$ stand for the cardinality of $ x_i$ and $n_\alpha=\prod_{i\in N(\alpha)} n_i$ be the cardinality of $\bfx_\alpha$. Then,
$$0 \le \bftheta^\top \bfb_\epsilon - \bftheta^\top \bfb^* \le \delta,$$
where
$$\delta = \epsilon\left(\sum_{\alpha} c_\alpha\ln n_\alpha + \sum_{i} c_i\ln n_i + \sum_i\sum_{\alpha\in N(i)} c_{i\alpha}\ln\frac{n_\alpha}{n_i}\right).$$
\end{proposition}
{\bf Proof:\ } The sets of beliefs $\bfb^*, \bfb_\epsilon$ are both in the local polytope $\LG$ whereas the beliefs $\bfb^*$ are the optimal ones with respect to the original linear program eqn.~\ref{eq:LP}, therefore  $\bftheta^\top \bfb^* \le \bftheta^\top \bfb_\epsilon$. On the other hand the beliefs $\bfb_\epsilon$ are optimal for the perturbed program eqn.~\ref{eq:LPe}, hence $\bftheta^\top \bfb_\epsilon \le \bftheta^\top \bfb^* + \epsilon(\tilde{H}(\bfb_\epsilon) - \tilde{H}(\bfb^*))$ where $\tilde H(\bfb)$ is described in eqn.~\ref{eq:H}.

Using Jensen's inequality we obtain:
$$H(\bfb_i)=\sum_{ x_i} b_i(x_i)\ln\frac{1}{b_i(x_i)}\le \ln \sum_{ x_i} \frac{b_i(x_i)}{b_i(x_i)} = \ln n_i,$$
and likewise $H(\bfb_\alpha)\le \ln n_\alpha$. Substituting in eqn.~\ref{eq:H} and noting that $\tilde{H}(\bfb^*)\ge 0$ we obtain:
$$\tilde{H}(\bfb_\epsilon) - \tilde{H}(\bfb^*) \le \sum_{\alpha} c_\alpha\ln n_\alpha + \sum_{i} c_i\ln n_i + \sum_{i,\alpha} c_{i\alpha}\ln\frac{n_\alpha}{n_i}.$$
\eop \vspace{0.2cm}

As a result, in the ideal world, one could generate the solution $\bfb_\epsilon$ arbitrarily close to the relaxed LP solution $\bfb^*$. There are, however, numerical accuracy limitations which in practice limit the size of $\epsilon\ge \epsilon_0>0$. The assumption in Proposition~\ref{prop:bound} is that the output $\bfb_\epsilon^{n.p.}$ of the norm-product algorithm, as defined in eqns.~\ref{eq:bi2},\ref{eq:balpha2}, is equal to $\bfb_\epsilon$ the solution to the $\epsilon$-perturbed LP-relaxation eqn.~\ref{eq:LPe}. This is indeed true when $\epsilon>0$ but not when $\epsilon=0$. As we shall see in more details in the next section, the norm-product algorithm is guaranteed to converge when $\epsilon=0$ but not necessarily to the minimal primal value.
Therefore, from a numerical perspective there exists $\epsilon_0$ such that when $\epsilon<\epsilon_0$ the underlying assumption $\bfb_{\epsilon}^{n.p.}=\bfb_{\epsilon}$ ceases to hold. Moreover, the value of $\epsilon_0$ depends on the graph structure and the potential functions $\psi_\alpha$ and therefore is unlikely to have a simple and useful form.

%Finally, if the norm-product algorithm produces beliefs without ties, i.e., $b_i(x_i)$ as produced by the algorithm has a single maximal entry for each variable node $i$, then the LP-relaxed solution $(b_i(x_i))^{1/\epsilon}$ approaches an integer value as $\epsilon\rightarrow 0$, and that in turn is the MAP.
%\begin{proposition}
%When $\epsilon\rightarrow 0$ and $-\tilde H$ is strictly convex, then when the norm-product algorithm converges to a solution $b_i(x_i)$ without ties, then the MAP solution follows from the index of maximal value of $b_i(x_i)$, $i=1,...,n$.
%\end{proposition}
%In reality, one often lacks a clear-cut determination of whether the solution is without ties. What we do in practice is compare $\bftheta^\top \bfb_\epsilon$ to $\bftheta^\top \hat \bfb_\epsilon$ where $\hat \bfb_\epsilon$ is the integer solution assignment. If the two are equal then we are guaranteed to have the MAP. 

\subsection{Convex-max-product Algorithm}
\label{sec:cmp}

We saw that for the setting of $\epsilon=0$ and when $\tilde H$ equals the Bethe entropy approximation then the norm-product becomes the max-product algorithm. We now explore the convex-free-energy setting $c_\alpha>0, c_i,c_{i\alpha}\ge 0$ while $\epsilon=0$ and refer to the resulting family of algorithms as  {\it convex-max-product\/} summarized in Fig. \ref{fig:convex-max-prod}.

Note that when $c_{i\alpha}=0$ we obtain a much simpler form of the algorithm where the message $n_{i\rightarrow\alpha}(\bfx_\alpha)$ depends only on $ x_i$ described in eqn.~\ref{eq:nia-short}:
\begin{algorithm}[Convex-Max-Product when $c_{i\alpha}=0$]
\label{alg:cmp-short}
Repeat until convergence:
\begin{enumerate}
\item For $i=1,...n$ and for all $\alpha\in N(i)$ do:
\beas
m_{\alpha\rightarrow i}(x_i)&=&\max_{ \bfx_\alpha\setminus x_i}\left\{\psi_\alpha(\bfx_\alpha)\prod_{j\in N(\alpha)\setminus i}n_{j\rightarrow\alpha}(x_j)\right\}\\
n_{i\rightarrow\alpha}(x_i)&\propto&  \frac{\displaystyle \left(\phi_i(x_i) \prod_{\beta\in N(i)}m_{\beta\rightarrow i}(x_i) \right)^{c_\alpha / \hat c_i} }{m_{\alpha\rightarrow i}(x_i)}
\eeas
\end{enumerate}
\end{algorithm}
The desired output vector $b_i(x_i)$ is recovered from computing the vector $\phi_i^{1/\hat c_i}(x_i) \prod_{\alpha\in N(i)}m^{1/\hat c_i}_{\alpha\rightarrow i}(x_i)$ as follows. If there are no ties, $b_i(x_i)$ is determined by setting the highest value to 1 and all remaining entries to 0. 
If the highest value of the vector is shared among $r_i>1$ entries, i.e., there exist ties, then those entries receive the value $1/r_i$. If there are no ties, i.e., $r_i=1$ for $i=1,...,n$, then the result is the MAP solution.% (see Appendix~\ref{app:MAP}).

The setting $\epsilon=0$ raises two issues (i) if the algorithm converges, can one obtain from them the optimal LP-relaxation solution?, and (ii) is there a convergence guarantee of the convex-max-product family? The answer to the first question is generally negative. In a nutshell, the primal function $f_{\epsilon=0}$ is convex but no longer strictly convex and therefore the dual function is no longer differentiable. A dual ascent approach on a non-differentiable dual function can get stuck at "corners". The implication of getting stuck at a corner of the energy landscape is that the recovered primal solution $\bfb_{\epsilon=0}$ might not correspond to the lowest primal energy and furthermore might not satisfy the marginal consistency constraints. More details can be found in Appendix~\ref{app:non-strict}.

We consider now the  the second question of whether the dual ascent creates a converging sequence? The answer is positive, i.e., the convex-max-product algorithm is convergent (unlike max-product on general graphs).  \\ 
\begin{theorem}[Convergence, Convex-max-product]
\label{theorem:cmp}
The norm-product algorithm with the parameter setting of $\epsilon=0$ and $c_\alpha>0, c_i,c_{i\alpha}\ge 0$ is convergent.
\end{theorem}
{\bf Proof:\ } Let $q_\epsilon(\bflambda_1,...,\bflambda_n)$ represent the conjugate dual eqn.~\ref{eq:dual1}:
$$q_\epsilon(\bflambda_1,...,\bflambda_n)=-f_\epsilon^*(-\sum_i \bflambda_i) - \sum_{i=1}^n h_{\epsilon,i}^*(\bflambda_i),$$
and let $q_0(\bflambda_1,...,\bflambda_n)$ be the limit of $q_\epsilon$ as $\epsilon\rightarrow 0$.  The explicit form of the conjugate duals $f_\epsilon^*$ and $h_{\epsilon,i}^*$ are:
\bea
f_\epsilon^*(\bflambda) \hspace{-0.2cm} &=& \hspace{-0.2cm} \sum_{\alpha}\ln \| \psi_\alpha(\bfx_\alpha)\exp(\lambda_\alpha(\bfx_\alpha))\|_{1/\epsilon c_\alpha}\\
h_{\epsilon,i}^*(\bflambda) \hspace{-0.2cm} &=&  \hspace{-0.2cm} \ln \left\| \phi_i(x_i) \hspace{-0.1cm}  \prod_{\alpha\in N(i)} \lp_{\tiny{ \bfx_\alpha\setminus x_i}}\exp(\lambda_\alpha(\bfx_\alpha))\|_{1/\epsilon c_{i\alpha}}\right\|_{1/\epsilon c_i},
\eea
where $\|_{\bfx_\alpha \setminus x_i} z(\bfx_\alpha) \|^p_p = \sum_{\bfx_\alpha \setminus x_i} |z_\alpha(\bfx_\alpha)|^p$.  The functions $f_0^*\stackrel{def}{=} f_{\epsilon\rightarrow 0}^*$ and $h_{0,i}^*\stackrel{def}{=} h_{\epsilon\rightarrow 0,i}^*$ are well defined and thus,
$$q_0(\lambda_1,...,\lambda_n)=-f_0^*(-\sum_i \bflambda_i) - \sum_{i=1}^n h_{0,i}^*(\bflambda_i),$$
is well defined as well. By definition of the block ascent scheme, let $\bflambda_{\epsilon,i}\in \argmax_{\bflambda_i}q_\epsilon(\bflambda_1,...,\bflambda_n)$. We note that $\bflambda_{0,i}=\lim_{\epsilon\rightarrow 0}\bflambda_{\epsilon,i}$ is well defined because $\epsilon$ appears as a norm in the definition of the message $n_{i\rightarrow\alpha}$.

We use the shorthand $q_\epsilon(\bflambda_{\epsilon,i})$ instead of 
$q_\epsilon(\bflambda_1,...,\bflambda_{i-1}, \bflambda_{\epsilon,i}, \bflambda_{i+1},...,\bflambda_n)$.
We wish to show that $\bflambda_{0,i}\in \argmax_{\bflambda_i}q_0(\bflambda_1,...,\bflambda_n)$. 

Assume to the contrary that $\bflambda_{0,i}\not\in \argmax_{\bflambda_i}q_0(\cdot)$ and let instead $\hat\bflambda_{0,i}\in \argmax_{\bflambda_i}q_0(\cdot)$, thus making $q_0(\hat\bflambda_{0,i})> q_0(\bflambda_{0,i})$. Since $q_0=\lim_{\epsilon\rightarrow 0} q_\epsilon$, there exists $\epsilon_0$ such that for all $\epsilon \le \epsilon_0$ we have $q_\epsilon(\hat\bflambda_{0,i})> q_0(\bflambda_{0,i})$ as well. Likewise, using the limit argument on the right-hand side, $q_\epsilon(\hat\bflambda_{0,i})> q_\epsilon(\bflambda_{0,i})$. Finally, since $\bflambda_{0,i}=\lim_{\epsilon\rightarrow 0}\bflambda_{\epsilon,i}$, and $q_\epsilon$ is continuous, we have $q_\epsilon(\hat\bflambda_{0,i})> q_\epsilon(\bflambda_{\epsilon,i})$ which contradicts the fact that $\bflambda_{\epsilon,i}\in \argmax_{\bflambda_i}q_\epsilon(\cdot)$. \eop \vspace{0.2cm} 

We conclude that the convex-max-product, unlike max-product, is convergence guaranteed, since it iteratively improves the dual objective which is bounded by the primal objective. The convex max-product is guaranteed to recover the MAP assignment if its beliefs are integral. However, in many cases we can use the rounding scheme for the max-product type algorithms which guarantees the MAP  if the beliefs recovered from the messages are without ties \cite{Weiss-uai07}.

\section{Experiments}

In our experiments we first evaluated the quality of the max-product type algorithms for solving a linear program with pairwise interactions and binary variables $$\min_{\bfb_i, \bfb_{i,j} \in \LG} \sum_{i,x_i \in \{0,1\}} \hspace{-0.2cm} \theta_i(x_i) b_i(x_i) + \hspace{-0.9cm} \sum_{(i,j) \in E, x_i,x_j \in \{0,1\}} \hspace{-0.9cm} \theta_{i,j}(x_i,x_j)b_{i,j}(x_i,x_j)$$ The max-product type algorithms differ from each other by their approximated entropy coefficients $c_\alpha, c_i, c_{i \alpha}$, but since the linear program has no entropy terms, all these algorithms aim at producing the same result. We distinguish between three families of max-product type algorithms: 
\begin{itemize}
\item The first family corresponds to non-concave entropy approximation, such as the Bethe free energy whose coefficients $c_\alpha=0, c_i=1-d_i$ and $c_{i \alpha}=0$ produce the max-product algorithm. These algorithms are not guaranteed to converge and even if they converge there are no guarantees on their solution. 
\item The second family corresponds to concave entropy approximations with positive $c_\alpha$, negative $c_i$ and $c_{i \alpha}=0$. The notable member of this family is the max-TRBP algorithm \cite{Wainwright-05map}, whose $c_\alpha$ is the weighted number of spanning trees which pass through the edge $\alpha$ and $c_i = 1-\sum_{\alpha \in N(i)} c_\alpha$. These max-product type algorithms are not guaranteed to converge, but whenever they converge one can extract an optimal solution for a pairwise linear program with binary variables, cf. \cite{Kolmogorov-uai05} theorem 4 and \cite{Meltzer-iccv05} corollary 2. 
\item The third family corresponds to concave entropy approximation with $c_\alpha, c_i, c_{i \alpha} \ge 0$. These convex-max-product algorithms are guaranteed to converge to the global optimum for a pairwise linear program with binary variables, cf. \cite{Meltzer-iccv05} corollary 2 and \cite{Globerson-nips07} proposition 3.  
\end{itemize}

We used the  implementation of the max-product type algorithm described in Algorithm~\ref{alg:cmp}, while each algorithm differs in its appropriate $c_\alpha, c_i, c_{i \alpha}$. To evaluate the performance of the algorithms we generated 100 samples of $10 \times 10$ grids, where $\theta_i$ and $\theta_{i,j}$ were sampled from zero mean Gaussians with standard deviation of one. We set the local evidence according to $\theta_i(x_i)=  \theta_i (-1)^{x_i}$, and for the pairwise interactions $\theta_{i,j}(x_i,x_j)$ we set the value $\theta_{i,j}$ on their diagonal and $-\theta_{i,j}$ on their off-diagonal. 

First we investigated the convergence properties of three representatives of the max-product families described above: The max-product algorithm, the max-TRBP described in \cite{Wainwright-05map}, and the convex max-product with the same tree-reweighted free energy, represented by $c_\alpha, c_i, c_{i \alpha} \ge 0$ as described in Appendix \ref{app:confree}. The convergence criterion for the max-product and max-TRBP algorithms was measured with respect to change in their messages, whereas the convergence criterion for the convex-max-product was measured with respect to change in its dual function. The max-product algorithm converged for $25\%$ of the runs, the max-TRBP converged for $90\%$ of the runs, and as expected from Theorem \ref{theorem:cmp} the convex-max-product always converged. However, the convex max-product was slower than max-TRBP, while we measured the primal values obtained by both algorithms during their runs. Over the runs the max-TRBP converged in average number of $430$ iterations compared to an average of $6400$ of the convex-max-product with tree-reweighted parameters.    

Next we compared the run-time of three representatives of the converging max-product: The convex-max-product with tree-reweighted free energy, the NMPLP of \cite{Globerson-nips07} and the convex-max-product with $c_\alpha=1, c_i=0, c_{i \alpha}=0$ which was referred as "trivial convex-max-product" by \cite{Weiss-uai07}. We measured their convergence with respect to the change in their dual objective: The NMPLP converged in average number of $200$ iterations, the trivial convex-max-product converged in average of $260$ iterations, and the convex-max-product with tree-reweighted free energy converged in average of $6400$ iterations. 

To conclude, for linear programs with pairwise interactions and binary variables the convex-max-product algorithms improve upon previous max-product type algorithms: They are guaranteed to converge to the global optimum. However the convex-max-product algorithms differ from each other in their memory requirements and run-time. Among those algorithms, the ones with $c_{i \alpha}=0$ requires less memory, as their messages $n_{i \rightarrow \alpha}$ depend only on $x_i$, and  have a faster run-time. 

The norm-product family of algorithms can also solve linear program using the perturbation method for a small value of $\epsilon$, as described in Proposition \ref{prop:bound}. However the convex-max-product algorithms are computationally more efficient, and guaranteed converge to the global optimum of linear program with pairwise interactions and binary variables. Therefore we evaluate the convex-norm-product type algorithms over linear programs with non-binary variables $$\min_{b_i, b_{i,j} \in \LG} \sum_{i,x_i \in \{1,2,3\}} \hspace{-0.4cm} \theta_i(x_i) b_i(x_i) + \hspace{-1.1cm} \sum_{(i,j) \in E, x_i,x_j \in \{1,2,3\}} \hspace{-1.1cm} \theta_{i,j}(x_i,x_j)b_{i,j}(x_i,x_j)$$  For these programs the convex-norm-product algorithms are guaranteed to converge to the global optimum, whereas the convex-max-product can converge to non-optimal stationary point. To evaluate the performance of the convex-norm-product we generated 100 samples of $10 \times 10$ grid where $\theta_i(x_i)$ and $\theta_{i,j}$ were sampled from zero mean Gaussians with standard deviation of one, and $\theta_{i,j}(x_i,x_j)$ were given the value $\theta_{i,j}$ on their diagonal and $-\theta_{i,j}$ on their off-diagonal. 

We measured how often the convex-max-product algorithm converges to non-optimal stationary points, comparing to the convex-norm-product which always achieves its optimum as described in Claim \ref{claim:convergence}. To indicate these events we compared the dual value of the linear program, which is evaluated by the convex-max-product stationary messages and by the convex-norm-product messages, setting $\epsilon=0.001$ and $c_\alpha=1, c_i=0, c_{i \alpha}=0$. For $60\%$ of the runs, the dual values attained by the convex-max-product and the convex-norm-product were $0.01$ close to each other, indicating both algorithms reached the maximal dual value. On the other hand, for $25\%$ of the runs the dual value of the linear program attained by the convex-max-product messages was $0.1$ lower than the one attained by the convex-norm-product messages, indicating the convex-max-product reached a non-maximal dual value. This fact has important practical implications: Only from the dual optimal solution one can recover the optimal beliefs that solve the primal linear program, while non-optimal dual messages always relate to non-consistent beliefs. In particular for the $25\%$ of the runs the convex-max-product did not produce beliefs which agree on their marginal probabilities, whereas the convex-norm-product always recover beliefs which satisfy the primal linear program constraints.  

In our experiments we also evaluated the sum-product type algorithms for approximating the marginal probabilities of distribution $p(\bfx)$ of the form $$p(\bfx) \propto \exp\left(\sum_{i,x_i} \theta_i(x_i) +  \sum_{\alpha, \bfx_\alpha} \theta_\alpha(\bfx_\alpha) \right)$$ The variational framework for approximating marginal probabilities, described in Section \ref{sec:var-inference}, suggests that the approximated entropy term affects the quality of the approximated marginal probabilities. Although we do not have a theoretical guarantee for setting the best approximation, in these experiments we show how the different approximations behave in practice. We consider two types of free energy approximations:
\begin{itemize}
\item Non-convex free energy approximations, represented by the Bethe approximation which corresponds to $c_\alpha=1, c_i=1-d_i, c_{i \alpha}=0$. The sum-product algorithm aims at finding a local minimum for the Bethe free energy approximation, but it is not guaranteed to converge. In cases where it does not converge we used the double loop algorithm \cite{Heskes06} in libDAI \cite{libdai}, which is guaranteed to converge to a stationary point of the Bethe free energy. 
\item Free energy approximations which are convex in the intersection of the marginalization constraints. These approximations are appealing since their stationary points are their global minimum. We address the tree-reweighted free energy approximations whose $c_\alpha, c_i, c_{i \alpha}$ correspond to spanning trees in the graph, and also to $L_2$ convex free energy approximation heuristic described in Appendix \ref{app:confree}. We note that whenever $c_\alpha, c_i, c_{i \alpha} \ge 0$ the corresponding convex-sum-product algorithms are guaranteed to converge to the global optimum. 
\end{itemize}

We used the implementation of the sum-product type algorithm described in Algorithm~\ref{alg:csp}, while each algorithm differs in its appropriate $c_\alpha, c_i, c_{i \alpha}$. Following \cite{Wainwright-05upper} We generated 100 samples of $10 \times 10$ grids with binary variables $x_i \in \{0,1\}$, where $\theta_i$ were uniformly chosen from the interval $[-0.05,0.05]$, and $\theta_{i,j}$ were either chosen uniformly from the {\em attractive} interval $[0,\omega]$ or the {\em mixed} interval $[-\omega, \omega]$. We ran the simulations with edge strength $\omega$ ranging from $0$ to $2$. We set the local evidence to $\theta_i(x_i)=  \theta_i (-1)^{x_i}$, and for the pairwise interactions $\theta_{i,j}(x_i,x_j)$ we set the value $\theta_{i,j}$ on their diagonal and $-\theta_{i,j}$ on their off-diagonal. 

\begin{figure}
\centerline{
\begin{tabular}[t]{c}
\includegraphics[width=7cm]{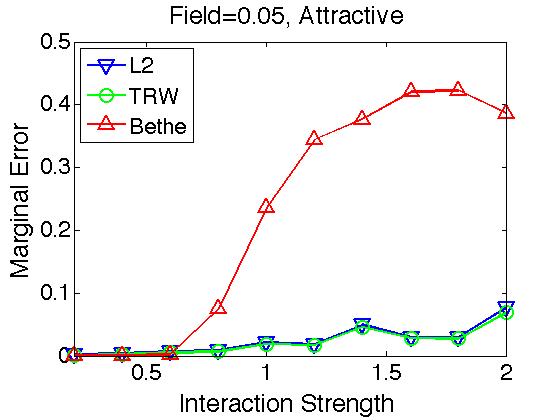} \\
\includegraphics[width=7cm]{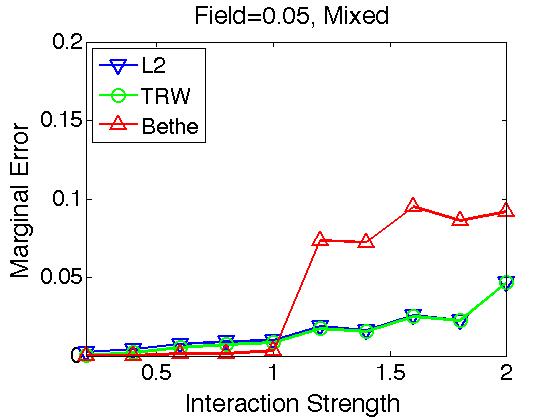} \\
\end{tabular}
}
\caption{\it\small
Comparison of error in marginal probabilities, estimated by Bethe free energy, tree-reweighted free energy and $L_2$ convex free energy described in Appendix~\ref{app:confree}. We computed the Bethe approximation by applying the sum-product when converged, and the double-loop algorithm otherwise. The other free energy approximations are convex and the convex-sum-product algorithm is guaranteed to converge to their optimum. The graphs present the average error over $100$ random trials}
\label{fig:grid}
\end{figure}

We compared to true marginal probabilities with the approximated marginal probabilities recovered from the Bethe free energy approximation, tree-rewighted free energy approximation, and the $L_2$ convex-free-energy heuristic. Fig. \ref{fig:grid} shows the average $L_1$ error in the marginal probabilities $\frac{1}{100} \sum_i |p^{(alg)}(x_i=1) - p^{(true)}(x_i=1)|$. 

We conclude from this experiment that the convex approximations are better than the Bethe approximation for the attractive settings, when $\theta_{ij} \ge 0$. However, the Bethe approximation is slightly better in the mixed settings for $\omega < 1$ and considerably  worse for $\omega > 1$. Moreover, in the mixed settings the sum-product did not converge for $\omega > 1$ and we used the double-loop algorithm instead which is computationally more expensive. We also conclude that the $L_2$ convex free energy settings produce comparable results to tree-rewiehted free energy for grids. 

\begin{figure}
\centerline{
\begin{tabular}[t]{c}
\includegraphics[width=7cm]{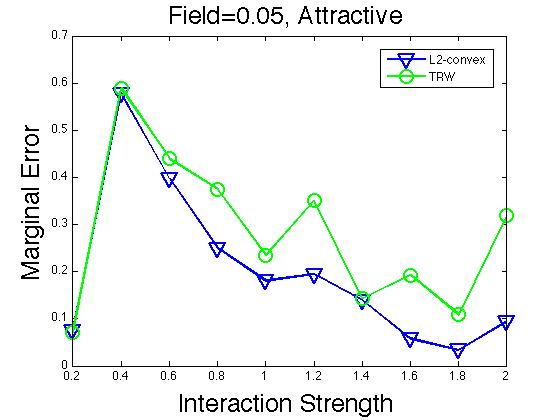} \\
\end{tabular}
}
\caption{\it\small
Comparison of error in marginal probabilities on a complete graph, estimated by tree-reweighted free energy approximation and $L_2$ convex free energy approximation. The graphs present the average error over $100$ random trials}
\label{fig:complete}
\end{figure}

We also compared the tree-reweighted and $L_2$ convex free energy approximated marginal probabilities on the complete graph, i.e. every two vertices are connected with an edge. We generated 100 samples of complete graphs with $10$ vertices with binary variables, where $\theta_i$ were uniformly chosen from the interval $[-0.05,0.05]$, and $\theta_{i,j}$ were chosen uniformly from the interval $[0,\omega]$, for $\omega$ ranging from $0$ to $2$. Fig. \ref{fig:complete} shows the average $L_1$ error in marginal probability, suggesting that in the case of complete graph, whose structure is far from a tree, the $L_2$ convex approximation heuristic is better than the tree-reweighted approximation for marginal probabilities estimation. 

Generally, the same convex free energy can be represented by different coefficients $c_\alpha, c_i, c_{i \alpha}$. In particular, the tree-reweighted free energy can be described by positive $c_\alpha$, which correspond to the weighted number of spanning trees that go through the edges $\alpha$, and negative $c_i = 1-\sum_{\alpha \in N(i)} c_\alpha$ and $c_{i \alpha}=0$. However, the same tree-rewieghted free energy can be represented by $c_\alpha, c_i, c_{i \alpha} \ge 0$, as explained in Appendix \ref{app:confree}. These representations affect their corresponding sum-product type algorithms: The first representation corresponds to the sum-TRBP algorithm which is not guaranteed to converge, whereas the second representation corresponds to the convex-sum-product which is guaranteed to converge. However, the convex-sum-product was slower than sum-TRBP, while we measured the primal values obtained by both algorithms during their runs. Similar results were reported in \cite{Globerson-UAI07}.
    
\begin{figure}
\centerline{
\begin{tabular}{c}
\includegraphics[width=7cm]{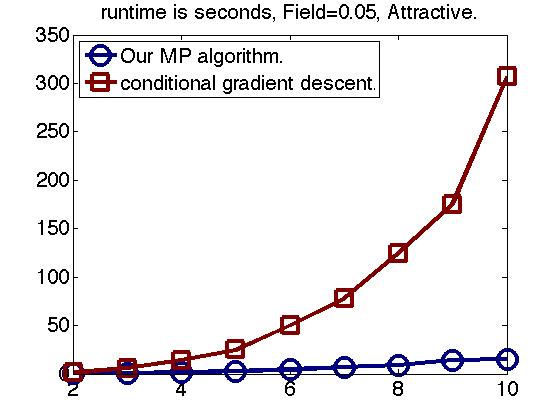} \\
\end{tabular}
}
\caption{\it\small
Run-time (in seconds) comparisons of convex-sum-product against a conditional gradient descent solver (running on convex-$L_2$ free energy). The algorithms were applied to $n\times n$ grids with $n=2,3...,10$.  Mean is shown for $10$ random trials.}
\label{fig:runtime}
\end{figure}

Fig. \ref{fig:runtime} compares the running time of the convex-sum-product algorithm with a general convex solver performing conditional gradient descent on the primal energy function \cite{Bertesekas03} which uses linear programming to find feasible search directions. We ran the algorithms on $n\times n$ grids where $n=2,3,...,10$. The stopping criteria for all algorithms was the same and based on a primal energy difference of $10^{-5}$. For a $10\times 10$ grid, for instance, the general convex solver was slower by a factor of 20 (e.g., $306$ seconds compared to $15.2$). For a $2 \times 2$ grid, on the other hand, convex-sum-product  took $0.15$ seconds compared $1.41$ seconds for the general convex solver. We conclude that the sum-product type algorithms converge faster than a general convex solver, since they exploit the structure of the graph. 

\section{Discussion}

We have presented a single unified message-passing framework for approximate inference covering both marginal probabilities estimation and the MAP assignment problem through LP-relaxation. We took a general perspective on the existing BP and TRBP algorithms and noted that all are reductions from the basic optimization formula of $f + \sum_i h_i$ where the function $f$ is an extended-valued, strictly convex but non-smooth and the functions $h_i$ are extended-valued functions (not necessarily convex). We used tools from convex duality to present the "primal-dual ascent" algorithm which is an extension of the Bregman successive projection scheme. Most of the details of this part of the paper was pushed to Appendix~\ref{app:dykstra} in order to reduce the overall technical load for the main-body presentation.

We then mapped the fractional-free-energy variational principal for approximate inference onto the optimization structure $f + \sum_i h_i$ and introduced the "norm-product" message-passing algorithm. Special cases of the norm-product include sum-product and max-product (BP algorithms), TRBP and NMPLP algorithms. When the fractional-free-energy is set to be convex (convex-free-energy) the norm-product is globally convergent for the estimation of marginal probabilities (the convex-sum-product branch corresponding to $\epsilon=1$) and for approximating the LP-relaxation ($\epsilon\rightarrow 0$). We have also introduced another branch of the norm-product which arises as the "zero-temperature" of the convex-free-energy ($\epsilon=0$) which we referred to as the convex-max-product. The convex-max-product is a convergent solver to the LP-relaxation (unlike max-product) but is not guaranteed to reach the global optimum. 
%It is interesting to note that while NMPLP was shown to be a special case of the norm product algorithm, it is not a member of the convex-max-product family. The reduction to NMPLP was under the setting $\epsilon=0$ but with negative values for $c_i$ whereas the convex-max-product family is defined for non-negative $c_i$. Given that NMPLP is convergent, we can conclude that the conditions $c_\alpha>0, c_i,c_{i\alpha}\ge 0$ are sufficient  for obtaining a convergent MAP approximation algorithm but are not necessary conditions.

As a general statement, the convex-free-energies provide a way for obtaining approximate inference over general graphs. There are two main issues in this regard: the first is how to obtain a guaranteed globally convergent message-passing algorithm for the general class of convex free energies, and secondly,  how to tune the energy parameters $c_i,c_{i \alpha},c_\alpha$ to a specific graph?

As for the first issue, we have provided a complete treatment which also encompasses the existing BP and TRBP algorithms (though they do not arise from a convex-free-energy but from a non-convex fractional-free-energy). 
As for the second issue, we provided a simple algorithm for converting the conventional TRW-free-energy settings to  the convex-free-energy framework and  have also proposed a heuristic  principle where among all admissible parameters we choose the one most closest to the Bethe free energy (Appendix~\ref{app:confree}). Empirical results show that for certain graphs, like a grid, we obtain very close marginal probability estimation results to those obtained by the TRW free energy. For complete graphs we obtain a very different free energy from TRW and superior accuracy of marginal probability estimation.  The results suggest that our heuristic for setting up the convex free energy satisfies what we were after, i.e., to get approximations similar to BP but in guaranteed (globally) convergent framework. 

In this work we limited the scope to factor graphs where the neighborhoods of every pair of factor nodes have at most a single intersection to give a clear description of the mathematical details presented in this work. However, the techniques presented here can also be used as a basis to a convex and non-convex generalized belief propagation \cite{Yedidia05}. Different algorithms were recently developed for tightening the LP-relaxation \cite{Sontag-nips08}, \cite{Sontag-uai08}, \cite{Komodakis-eccv08} using intersections of increasingly  larger clusters in order to recover the MAP assignment. We believe similar techniques can be applied to convex free energies in order to tighten the bound on the log-partition function. 
 
We did not discuss the parallel implementation of the norm-product algorithm, but as every message-passing algorithm it can be parallelized: One can distribute to the different parallel units an independent set of vertices, i.e. vertices which are not connected to each other in the graph. This mechanism preserves the convergence and optimal guarantees of the algorithm. The norm-product can also be made fully parallel, as it is a generalization of the belief propagation algorithm, but in this case convergence is no longer guaranteed. This can be fixed by methods described in \cite{Hazan-Shashua-uai08}. 
 
The convergence rate and the complexity analysis of the norm-product algorithm were not addressed in this work. Since the convex norm-product algorithm performs a dual block ascent it has a linear convergence rate, whenever $\epsilon ,c_\alpha,c_i > 0, c_{i \alpha} =0$ (cf. \cite{Luo-Tseng93}  Theorem 5.1), i.e. it achieves a $\delta$-optimal solution in $O(\log(1/\delta))$ steps. However, this notation does not capture the true complexity of the algorithm as $O(\log(1/\delta))$ depends on unknown constants which can be very large. For this purpose  complexity bound were recently introduced, where it was proved that the dual gradient ascent attains linear complexity, (cf. \cite{Nesterov04} Theorem 2.1.13, \cite{Tseng09} Theorem 5.1). Although the convex norm-product can be modified to achieve linear complexity its step size depends on $\epsilon,c_\alpha,c_i$ and the modified algorithm is inefficient compared to the convex norm-product. We believe this due to the fact that the convex norm-product finds the optimal dual assignment $\bflambda_i$ in each step, unlike the gradient methods. Generally, a complexity bound for block coordinate ascent algorithms such as the convex norm-product is an open problem. 

Future work is also required for obtaining a firmer theoretical understanding about how to set the concave entropy approximation, in order to guarantee a good approximation for the marginal probabilities. For example, how tight is the TRW-entropy bound, and whether one can find a family of trees which guarantees the best bound? Clearly, these theoretical guarantees must consider the potentials functions, since for every graph its TRW-entropy can be made arbitrary close to the true entropy for some potentials. %We conjecture that the entropy approximation is not important for the convex max-product algorithm, since it solve the LP-relaxation which is independent of the entropy. 

\appendices

\section{Mathematical Background on Conjugate Duality}
\label{app:bck}

We consider the n-dimensional Euclidean space $R^n$ and denote vectors in bold face, e.g. $\bfx \in R^n$. We start with a brief review of basic concepts of sets. A set $S$ is said to be {\em closed} if every of its limit points is contained the set. A set $S$ is called {\em open} if its complement $R^n \setminus S$ is closed. The {\em interior} of a set $S$, denoted by $int(S)$, is the largest open set contained in $S$. The closure of a set, $cl(S)$,  is the smallest closed set containing $S$. 
A point $\bfx$ is a {\it boundary\/} point of $S$ if $\bfx\in cl(S)$ and $\bfx\not\in int(S)$ or equivalently if every neighborhood of $\bfx$ contains at least one point of S and at least one point not of S.
A set $C$ is called {\em convex} if it contains the line-segment between any two points $\bfx$ and $\bfy$ in the set. That is, for every $0\le \lambda \le 1$ the point $\lambda \bfx + (1-\lambda)\bfy \in C$.

For our purposes, since we deal with low-dimensional sets placed in higher-dimensional spaces, we use the concept of {\it relative interior\/}  denoted by $ri(S)$ which, defined intuitively,  contains all points which are not on the "edge" of the set, relative to the smallest affine subspace in which this set lies. For example, for a convex set $C$, $\bfx\in ri(C)$ if and only if $\forall \bfy\in C$ there exists $\bfz\in C$ and $0<\lambda < 1$ such that $\bfx = \lambda\bfz + (1-\lambda)\bfy$.
 
The {\em graph} of a function $f(\bfx)$ is the curve $\{(\bfx, f(\bfx)) : \bfx \in R^n\}$, and define the {\em epigraph} of a function $f(\bfx)$, denoted by $epi(f)$, as the set above its graph, namely $\{(\bfx, r) : \bfx \in R^n, r \ge f(\bfx)\}$. A functions is called {\em closed} if its epigraph is a closed set. A function is said to be {\em convex} if its epigraph is a convex set. A function is called {\em strictly convex} if any line segment in its epigraph intersects with its relative interior. A twice differentiable function is convex if its matrix of second derivatives, called the {\em Hessian}, is positive semidefinite, and strictly convex if its Hessian is positive definite.

In this paper we work with functions that can take the value of infinity and as such are non-differentiable. Such functions are known as {\it extended-valued\/}: \vspace{0.2cm}
\begin{definition}[Extended-Valued, Proper]
\label{def:proper}
A function $f(\bfx)$ is said to be extended real-valued if $-\infty \le f(\bfx) \le \infty$. The effective domain of $f(\bfx)$ is denoted by $dom(f) = \{\bfx: f(\bfx)<\infty\}$. A function is said to be proper if $-\infty < f(\bfx) \le \infty$, and it obtains at least one finite value.
\end{definition} \vspace{0.2cm}
Proper functions typically arise when constraints are embedded into finite valued functions. For example, the {\it indicator function\/} associated with a convex set $C$ is defined by $\delta_C(\bfx)=0$ when $\bfx\in C$ and $\delta_C(\bfx)=\infty$ otherwise. A possible use of the indicator function is to constrain a finite valued function $\hat f$ with the set convex set $S$ to define a proper function $f=\hat f + \delta_S$. We define next the type of smoothness used throughout this paper:

\vspace{0.2cm}
\begin{definition}[Essentially Smooth]
\label{def:smooth}
Let $f$ be a proper and convex function differentiable throughout the non-empty set  $C=int(dom(f))$. Then $f$ is called {\em essentially smooth\/} if $\lim_{k \rightarrow \infty} \|\nabla f(\bfx_k)\| = \infty$ whenever $ \bfx_k$  is a sequence in C converging to a boundary point $\bfx$ in $C$. 
\end{definition} \vspace{0.2cm}
Necessary and sufficient conditions for a function to be essentially smooth are described in the following theorem: \\ 
\begin{theorem}[Legendre type]
\label{theorem:smooth}
A closed and proper convex function $f(\bfx)$ is essentially smooth if and only if it is differential in its interior $C = int(dom(f))$, i.e. $\partial f(\bfx) = \nabla f(\bfx)$ for every $\bfx \in C$, while $\partial f(\bfx) = \emptyset$ when $\bfx \not \in C$. If $f(\bfx)$ is also strictly convex on $C$ it is called a convex function of {\em Legendre type}, and its gradient mapping $\nabla f: C \rightarrow R^n$ is continuous and one-to-one, and $\nabla f^* = (\nabla f)^{-1}$.
\end{theorem} \vspace{0.1cm}
{\bf Proof:} \cite{Rockafellar70}, Theorem 26.1 and Theorem 26.5 \eop \vspace{0.2cm}

The sets
$\{\bfx : \bfa^\top\bfx\ge b\}$ and $\{\bfx : \bfa^\top\bfx\le b\}$, 
are called the {\it closed half-spaces\/} associated with the hyperplane $\{\bfx : \bfa^\top\bfx = b\}$. 
We say that two sets $C_1,C_2$ are {\it separated by a hyperplane\/} if each set lies in a different closed halfspace associated with the hyperplane. If a vector $\bar\bfx$ is a boundary point  of a set $C$, then a hyperplane that contains the singleton $\{\bar\bfx\}$ and one of its halfspaces contains $C$ is said to be {\it supporting $C$ at $\bar\bfx$}. In other words, a supporting hyperplane is a hyperplane that "just touches" the set $C$. If $C$ is a convex set then there exists a supporting hyperplane for every point on its boundary. Supporting hyperplanes play a role in the definition of the sub-gradient of a non-differentiable function. A vector $\bflambda$ is called a {\it subgradient\/} of a convex proper function $f$ at $\bfx$ if  
\be \forall \bfz \;\;\;\;  f(\bfz) \ge f(\bfx) + \bflambda^\top (\bfz-\bfx).\label{eq:subg}
\ee
This condition has a simple geometric meaning: it says that  the affine function $h(\bfz) = f(\bfx) + \bflambda^\top (\bfz-\bfx)$ is a (non-vertical) supporting hyperplane to the convex set epi(f) at the point $(\bfx,f(\bfx))$. Consequently, the set of subgradients $\bflambda$ at $\bfx$, called the {\it subdifferential of $f$ at $\bfx$} and is denoted by  $\partial f(\bfx)$,  consists of the supporting hyperplanes to the convex set epi(f) at the point $(\bfx,f(\bfx))$. When $f$ is differentiable at $\bfx$ then the supporting hyperplane is unique and $\partial f(\bfx) = \nabla f(\bfx)$. 
\vspace{0.2cm}
\begin{definition}
\label{def:sub-d}
The sub-differential of a function $f$ at a point $\bfx$ is denoted by $\partial f(\bfx)$ and consists of all the supporting hyperplanes of epi(f)  at the point $\bfx$, namely $$\partial f(\bfx) = \{\bflambda : \forall \bfz \;\;\; f(\bfz) \ge f(\bfx) + \bflambda^\top(\bfz-\bfx)\}$$
\end{definition} \vspace{0.2cm}
The following claim describes the sub-differential of the indicator function associated with affine sets (a useful result which will serve us later): \vspace{0.2cm}
\begin{claim}
\label{claim:dif-affine}
Let $A$ be $k \times n$ matrix and consider the affine set ${\cal B} = \{\bfx : A \bfx = \bfc\}$ and its indicator function
$$\delta_{\cal B}(\bfx) = \left\{
\begin{array}{cc} 0 & A \bfx = \bfc \\ \infty & otherwise \end{array}
\right. $$
Then $\partial \delta_{\cal B} = \{A^\top \bfsigma : \bfsigma \in R^k\}$.
\end{claim} \vspace{0.1cm}
{\bf Proof:}
This claim results as a special case of \cite{Bertesekas03} example 7.1.4. For the sake of clarity we provide a direct proof.  We describe the sub-differential $\partial \delta_{\cal B}(\bfx)$ for every point $\bfx$ in the domain of $\delta_{\cal B}$, i.e., $\delta_{\cal B}(\bfx)=0$.
To prove the direction $\partial \delta_{\cal B} \supseteq \{A^\top \bfsigma : \bfsigma \in R^k\}$ we must show that  $\delta_{\cal B}(\bfz) \ge \delta_{\cal B}(\bfx) + \bfsigma^\top A (\bfz-\bfx)$ for every $\bfz$. For every $\bfz$ satisfying $A \bfz = \bfc$ this relation holds since $\delta_{\cal B}(\bfz)=0$ and $A (\bfz-\bfx) = 0$. For every $\bfz$ with $A\bfz \ne \bfc$ this relation holds since $\delta_{\cal B}(\bfz)=\infty$.

To prove the other direction $\partial \delta_{\cal B} \subseteq \{A^\top \bfsigma : \bfsigma \in R^k\}$ we must show that  $\delta_{\cal B}(\bfz) \ge \delta_{\cal B}(\bfx) + \bflambda^\top(\bfz-\bfx)$ only if $\bflambda =  A^\top \bfsigma$ holds for every $\bfz$. First we note that the set $\{\bfz - \bfx : A \bfz =\bfc \}$ is orthogonal to $\{A^\top \bfsigma : \bfsigma \in R^k\}$, therefore  if we assume on the contrary that $\bflambda \ne A^\top \bfsigma$ there must be a vector $\bfz_0-\bfx$ with non-vanishing angle with $\bflambda$, namely $\bflambda^\top (\bfz_0 - \bfx) > 0$ therefore Definition \ref{def:sub-d} does not hold for $\bflambda$. \eop \vspace{0.2cm} 

\begin{claim}
\label{claim:B}
Consider a function $f$ whose domain is contained in the affine set ${\cal B} = \{\bfx: A\bfx = \bfc\}$. Then whenever $\bflambda \in \partial f(\bfx)$ there holds $(\bflambda + A^\top \bfsigma) \in \partial f(\bfx)$ for every $\bfsigma$.
\end{claim} \vspace{0.1cm}
{\bf Proof:} $f(\bfx)$ can be equivalently written as $f(\bfx) + \delta_{\cal B}(\bfx)$ where $\delta_{\cal B}$ is the indicator functions of the affine set ${\cal B}$, therefore $\partial f = \partial (f + \delta_{\cal B})$. From the linearity of the sub-differential, cf. \cite{Bertesekas03} Theorem 4.2.4, there holds $\partial (f(\bfx) + \delta_{\cal B}(\bfx)) = \partial f(\bfx) + \partial \delta_{\cal B}(\bfx)$ and the claim follows since $\bflambda \in \partial f(\bfx)$ by assumptions, and $A^\top \bfsigma \in \partial \delta_{\cal B}(\bfx)$ from Claim \ref{claim:dif-affine}. \eop \vspace{0.2cm} 

A supporting hyperplane at $\bfx$ with $\bflambda$-slope must satisfy Definition \ref{def:sub-d}, namely
$$
\forall \bfz \;\;\;\; f(\bfz) - \bflambda^\top \bfz \ge f(\bfx) - \bflambda^\top \bfx,
$$
therefore it must hold that the $\bflambda$-slope hyperplane supports the epigraph at $(\bfx, f(\bfx))$ where $\bfx \in \argmin \{f(\bfz) - \bflambda^\top \bfz\}$. 
This leads to the definition of the {\it conjugate\/} function: \vspace{0.2cm}
\begin{definition}
\label{def:conj}
The Fenchel-Legendre conjugate is:
$$f^*(\bflambda)=\max_{\bfx\in dom(f)} \{(\bflambda^\top\bfx - f(\bfx)\}.$$
\end{definition}\vspace{0.2cm}

The conjugate  $f^*(\bflambda)$  describes the offset of the $\bflambda$-hyperplane that supports the epigraph of $f$. Note that regardless of the structure of $f(\bfx)$ its conjugate function $f^*(\bflambda)$ is closed and convex, since it is the pointwise maximum of a collection of affine (closed) functions. Furthermore, if $f$ is convex then the conjugate of its conjugate returns back $f$, i.e., $f^{**}=f$ (cf. \cite{Bertesekas03}, Theorem 7.1.1). The following claim is a useful result which shall serve us later:
\vspace{0.2cm}
\begin{claim}
\label{claim:conj-t}
Let $g^*(\bflambda) = f^*(\bflambda - \bfmu)$, then $g(\bfx) = f(\bfx) + \bfmu^\top \bfx$
\end{claim}
\vspace{0.1cm}
{\bf Proof:}
The definition of the Fenchel-Legendre conjugate of $g(\bfx) = f(\bfx) + \bfmu^\top \bfx$ takes the form $g^*(\bflambda) = \max_x \{\bflambda^\top \bfx - \bfmu^\top \bfx - f(\bfx)\}$
which has the form in the claim since $\bflambda^\top \bfx - \bfmu^\top \bfx = (\bflambda - \bfmu)^\top \bfx$ \eop \vspace{0.2cm}

The convex conjugate plays an important role in duality. Consider contrained minimization under linear constraints $A\bfx=0$, i.e., $\bfa_1^\top \bfx = 0,...,\bfa_k^\top \bfx = 0$ with $\bfa_i^\top$ being the i'th row vector of $A$. The statement  about the existence of Lagrange multiplier for non-differentiable functions is described next:
\vspace{0.2cm}
\begin{theorem} {\bf (Lagrange multipliers)} \\
\label{theorem:Lagrange-c}
Let $f(\bfx)$ be a proper convex function and consider the convex program 
$$\min_{\bfx \in dom(f)} f(\bfx) \hspace{0.3cm} \mbox{subject to} \hspace{0.3cm} A\bfx=0. $$
Assume $ri(dom(f))$ intersect the linear constraints $A\bfx=0$ and that the optimal value of the program is finite. Then there exists Lagrange multipliers $\lambda^*_1,...,\lambda^*_k$ satisftying
\be
\label{eq:l-opt}
\bfx^* \in \argminn{\bfx \in dom(f)} \{f(\bfx) + \sum_i \lambda^*_i \bfa_i\}
\ee
\end{theorem}
{\bf Proof:} The assumptions 6.4.1 in \cite{Bertesekas03} hold in this case and following the Nonlinear Farkas lemma, as done in Theorem 6.4.2 in \cite{Bertesekas03}, completes the proof. \eop \vspace{0.2cm} 

The duality theorem using the conjugate $f^*$ is described below:
\vspace{0.1cm}
\begin{theorem}{\bf (Strong Duality)}
\label{theorem:Lagrange-d}
Let $f$ be a convex proper function and $ri(dom(f))$ intersects with the constraints $A\bfx=0$, and that the optimal value of the program is finite. The following form a primal-dual pair:
\bea
(primal) && \min_{\bfx \in domain(f)} f(\bfx) \hspace{0.3cm} \mbox{s.t.} \hspace{0.3cm} A\bfx=0 \label{eq:primal} \\
(dual) &&  \max_{\bflambda = \lambda_1,...,\lambda_k} -f^*(-A^\top \bflambda) \label{eq:dual}
\eea
Then there is no duality gap and there exists primal-dual optimal pair. Moreover, the vectors $(\bfx^*,\bflambda^*)$ form a primal-dual optimal pair  $f(\bfx^*) = -f^*(-A^\top \bflambda^*)$ if and only if
the following "algorithmic certificate" for optimality hold:
\begin{eqnarray*}
\hspace{-1cm} \bfx^* \in dom(f)  \hspace{0.8cm} && \mbox{(feasibility)} \\
\hspace{-1cm} \bfzero \in \partial \{ f(\bfx^*) + A^\top \bflambda^* \}  && \mbox{(optimality)}
\end{eqnarray*}
\end{theorem}
{\bf Proof:} The existence of primal-dual optimal pair follows from Theorem \ref{theorem:Lagrange-c}. The rest follows from \cite{Bertesekas03}, Theorem 6.2.5 \eop \vspace{0.2cm} 

Note that due to the linearity of the sub-differential $\partial (f + g) = \partial f + \partial g$, the optimality condition above is equivalent to $-A^\top \bflambda^* \in \partial f(\bfx^*)$.

To see the connection to Lagrangian duality, note that by definition of $f^*$ we have:
$$-f^*(-A^\top \bflambda^*)=\min_{\bfx\in dom(f)} \{f(\bfx)+A^\top \bflambda^*\},$$
which in turn means that the primal-dual pair $(\bfx^*,\bflambda^*)$ satisfy $\bfx^*\in \argmin_{\bfx} \{f(\bfx)+A^\top \bflambda^*\}$ where the right-hand side is the Lagrangian $L(\bfx,\bflambda)=f(\bfx)+A^\top \bflambda$ and the dual problem is $\max_{\bflambda}q(\bflambda)$ where $q(\bflambda)=\min_{\bfx} L(\bfx,\bflambda)$.

A proper convex function $f(\bfx)$ is {\em essentially strictly convex} if it is strictly convex on every convex subset in $dom(\partial f)$.  We note below that in order for the dual function to be smooth the primal must be strictly convex. A smooth dual is necessary for a dual ascent scheme (described later). \vspace{0.2cm}
\begin{theorem} {\bf (strict primal $\Longleftrightarrow$ smooth dual)} \\
\label{theorem:strict-smooth}
A closed proper convex function is essentially strictly convex if and only if its conjugate it essentially smooth.
\end{theorem} \vspace{0.1cm}
{\bf Proof:} \cite{Rockafellar70}, Theorem 26.3 \eop \vspace{0.2cm} 

We describe below two Fenchel duality theorems which are the functional form of the Lagrange duality where the constraints are implicit in the functions domains:\vspace{0.2cm}
\begin{theorem}{\bf Basic Fenchel Duality I} \\
\label{theorem:FD1}
Let $g(\bfx), h(\bfx)$ be proper closed and convex functions and $ri(dom(g))\cap ri(dom(h))\not=\emptyset$, and the value of the program is finite. The following are primal and dual programs:
\begin{eqnarray*}
&\mbox{Primal:}& \min_{\bfx} g(\bfx) + h(\bfx)\\
&\mbox{Dual:}& \max_{\bflambda} -g^*(-\bflambda) - h^*(\bflambda)
\end{eqnarray*}
Then there is no duality gap, and there exists primal-dual optimal pair. Moreover, the vectors $(\bfx^*,\bflambda^*)$ are primal-dual optimal pair if and only if $-\bflambda^* \in \partial g(\bfx^*)$ and $\bflambda^* \in \partial h(\bfx^*)$. Conversely, by reversing the roles of primal and dual, the vectors $(\bfx^*,\bflambda^*)$ are primal-dual optimal pair if and only if $\bfx^* \in \partial g^*(-\bflambda^*)$ and $\bfx^* \in \partial h^*(\bflambda^*)$. In particular, if $g(\bfx)$ is essentially strictly convex and $g^*(\bflambda)$ is finite, then the optimal $\bfx^*$ is determined by $\bfx^* \in \nabla g^*(-\bflambda^*)$.
 \end{theorem}
\vspace{0.1cm}
 {\bf Proof:} We reduce Fenchel duality to Lagrange duality in Theorem \ref{theorem:Lagrange-d}, where we consider a decomposed version of the primal function $f(\bfx_g,\bfx_h) = g(\bfx_g) + h(\bfx_h)$ subject to the linear consistency constraints $\bfx_g=\bfx_h$. Note that the vector equality constraint is composed from $m$ equality constraints where $m$ is the length of the vectors $\bfx_g$ and $\bfx_h$, therefore we expect to use Lagrange multipliers vector $\bflambda$ of length $n$. The Lagrangian $L(\bfx_g,\bfx_h,\bflambda)$ takes the form $g(\bfx_g) + h(\bfx_h) + \bflambda^\top (\bfx_g-\bfx_h)$ and using the conjugate notation in Definition. \ref{def:conj} the dual function $q(\bflambda) = \min_{\bfx_g,\bfx_h} L()$ takes the form in the theorem above. Following Theorem \ref{theorem:Lagrange-d} there exists primal-dual optimal pair which must satisfy the feasibility condition, i.e. $\bfx^*_g = \bfx_h^*$, and the optimality condition, namely $-\bflambda^* \in \partial g(\bfx_g^*)$ and $\bflambda^* \in \partial h(\bfx_h^*)$.  The theorem follows as the optimal $\bfx^*$ must equal $\bfx^*_g$ as well as $\bfx^*_h$. Reversing the roles of primal and dual are allowed by convexity whereby $g^{**}=g, h^{**}=h$. Furthermore, 
since $g^*(\bflambda)$ is finite Theorem \ref{theorem:strict-smooth} determines $g^*(\bflambda)$ to be smooth, and whenever $\bfx^* \in \partial g^*(-\bflambda^*)$ there must hold $\bfx^* \in \nabla g^*(-\bflambda^*)$. \eop \vspace{0.2cm}

The next theorem is generalizes the Fenchel duality theorem above:
\vspace{0.2cm}
\begin{theorem}{\bf Basic Fenchel Duality II} \\
\label{theorem:FD2}
Let $f(\bfx), h_1(\bfx),..., h_n(\bfx)$ be proper, closed and convex functions and $ri(dom(f))\cap ri(dom(h_i))\not=\emptyset$ and the optimal value of the program is finite. The following are primal and dual programs:
\begin{eqnarray}
&\mbox{Primal:}& \min_{\bfx} f(\bfx) + \sum_{i=1}^n  h_i(\bfx)\nonumber\\
&\mbox{Dual:}& \max_{\bflambda} \{-f^*(-\sum_{i=1}^n \bflambda_i) - \sum_{i=1}^n h_i^*(\bflambda) \}\label{eq:dual1}
\end{eqnarray}
Then there is no duality gap, and there exists prima-dual optimal pair. Moreover, the vectors $(\bfx^*,\bflambda^*_i)$ are primal-dual optimal pair if and only if $-\sum_{i=1}^n \bflambda_i^* \in \partial f(\bfx^*)$ and $\bflambda^*_i \in \partial h_i(\bfx^*)$. Also, if $f(\bfx)$ is essentially strictly convex and $f^*(\bflambda)$ is finite, then $\bfx^* = \nabla f^*(-\sum_i \bflambda_i^*)$.
 \end{theorem}
\vspace{0.1cm}
 {\bf Proof:} The proof closely follows the one of Theorem \ref{theorem:FD1} where we consider a decomposed version of the primal function $f(\bfx_f) + \sum_i h_i(\bfx_i)$ subject to the linear consistency constraints $\bfx_f = \bfx_i$. The Lagrangian $L()$ takes the form $f(\bfx_f) + \sum_i h_i(\bfx_i) + \sum_i \bflambda_i^\top (\bfx_f - \bfx_i)$ and the dual function $q(\bflambda_i)$ takes the form in the theorem above. Following the Lagrange duality in Theorem \ref{theorem:Lagrange-d} there exists primal-dual optimal pair which must be primal feasible, i.e., $\bfx^*_f =  \bfx_i^*$, and satisfy the optimality condition $-\sum_{i=1}^n \bflambda^* \in \partial f(\bfx_f^*)$ and $\bflambda_i^* \in \partial h_i( \bfx_i^*)$.  Whenever $f(\bfx)$ is essentially strictly convex and $f^*(\bflambda)$ is finite, repeating the primal-dual reversing argument of Theorem \ref{theorem:FD1} shows that $\bfx^* = \nabla f^*(-\sum_i \bflambda_i^*)$. \eop \vspace{0.2cm} 
 
Algorithmically, minimizing the primal program $f(\bfx) + \sum_{i=1}^n  h_i(\bfx)$ requires to take into account the domains of $f$ and $h_i$ simultaneously.  Therefore, it is algorithmically appealing to solve the primal program in a piece-meal fashion using dual block ascent, while iteratively improving a single vector $\bflambda_i$. This way one need to consider only sub-problems that consists of $f^*$ and a single $h_i^*$. After we recover the optimal $\bflambda^*_i$ one can recover efficiently the primal optimal $\bfx^*$ by using the smoothness of $f^*$ as describes in Theorem \ref{theorem:FD2}:
\vspace{0.2cm}
\begin{algorithm}[Dual Block Coordinate Ascent]
\label{alg:dba}
Initialize $\bflambda_1=\bfzero,...,\bflambda_n=\bfzero$.
\begin{enumerate}
\item Repeat until convergence:
\item For $i=1,...n$:
  \begin{enumerate}
  \item $\bfmu_i \leftarrow   \sum_{j \ne i} \bflambda_j$
  \item $\bflambda_i \leftarrow \argmaxx{\bflambda_i} \{-f^*(-\bflambda_i - \bfmu_i) - h_i^*(\bflambda_i) \}$
  \end{enumerate}
\end{enumerate}
Output $\bfx^* = \nabla f^*(-\sum_i \bflambda_i^*)$.
\end{algorithm}
\vspace{0.2cm}
The dual block ascent algorithm iteratively improves the dual objective therefore is guaranteed to converge. Whenever $f(\bfx)$ is strictly convex in its domain its conjugate is essentially smooth and the dual block ascent is guaranteed to converge to the global optimum, as formally described below:
\vspace{0.2cm}
\begin{theorem} {\bf (Dual Block Ascent)}
\label{theorem:dba}
Let $f,h_i$ be closed convex functions and assume the relative interior of their domains intersect.
In addition, assume $h_i$ are continuous over their domains and $f$ is strictly convex  over its domain and $f^*$ is finite.  Then, the dual block ascent algorithm converges to the dual and primal optimum.

In particular, if the dual sequence is bounded then every of its limit points is an optimal dual solution $\bflambda_1^*,..., \bflambda_n^*$. Also, consider the primal sequence generated by $\nabla f^*(-\sum_i \bflambda_i)$ computed from the dual sequence, then this primal sequence is bounded and its limit point is the optimal solution $\bfx^*$.
\end{theorem} \vspace{0.1cm}
{\bf Proof:} \cite{Luo-Tseng93}. \eop \vspace{0.2cm}

\section{The Primal-Dual Block Ascent Algorithm}
\label{app:dykstra}

We describe an algorithm for solving programs of the form $$f(\bfb) + \sum_i h_i (\bfb)$$ while solving sub-problems which consists of $f(\bfb)$ and a single function $h_i(\bfb)$. In our framework we include convex as well as non-convex optimization, but for now we describe the convex settings, and later describe the necessary conditions for this optimization scheme for non-convex programs. The dual block ascent method, described in Algorithm \ref{alg:dba} decomposes the optimization program to sub-problems which solve a dual function which requires the explicit computation of the conjugate functions $f^*(\bflambda)$ and $h_i^*(\bflambda)$ --- a task which is often algorithmically unattractive or unfeasible. Instead, 
one can recover $\bflambda_i$ in Algorithm \ref{alg:dba} by solving its primal program and using the primal-dual optimality condition in Theorem \ref{theorem:FD1} as follows.  Set $h(\bfb) \leftarrow h_i(\bfb)$ and $g(\bfb) \leftarrow  f(\bfb) + \bfb^\top \bfmu_i$ and recall Claim \ref{claim:conj-t} from which we obtain $g^*(-\bflambda_i) = f^*(-\bflambda_i - \bfmu_i)$, and solve the primal program:
\be
\label{eq:pdba}
\bfb^* = \argminn{\bfb \in dom(f) \cap dom(h_i)} \left\{ f(\bfb) + \bfb^\top \bfmu_i + h_i(\bfb) \right\}
\ee

If the pair of functions $f(\bfb)$ and $h_i(\bfb)$ satisfy the assumptions of Theorem \ref{theorem:FD1} then the functions $g(\bfb) \leftarrow  f(\bfb) + \bfb^\top \bfmu_i$ and $h_i(\bfb)$ satisfy these assumptions as well and, hence, $\bflambda_i$ can be recovered from the optimality conditions of Theorem \ref{theorem:FD1}:

\be
\label{eq:pdba-opt}
\bflambda^*_i \in \{ -\bfmu_i - \partial f(\bfb^*) \} \cap \partial h_i(\bfb^*)
\ee

Taken together, one obtains the primal form of the dual block ascent algorithm, in which one need not compute the conjugate functions:
\vspace{0.2cm}
\begin{algorithm}[Primal-Dual Vanilla]
\label{alg:pdba}
Let the functions $f(\bfb)$ and $h_i(\bfb)$ satisfy the conditions of Theorem~\ref{theorem:dba}. Initialize $\bflambda_1=\bfzero,...,\bflambda_n=\bfzero$.
\begin{enumerate}
\item Repeat until convergence:
\item For $i=1,...n$:
  \begin{enumerate}
  \item $\bfmu_i \leftarrow   \sum_{j \ne i} \bflambda_j$
  \item $\bfb^* \leftarrow  \argminn{ \bfb \in dom(f) \cap dom(h_i)}\left\{f(\bfb) + h_i(\bfb) + \bfb^\top \bfmu_i \right\}$.
  \item Recover $\bflambda_i \in \{ -\bfmu_i - \partial f(\bfb^*) \} \cap \partial h_i(\bfb^*)$
  \end{enumerate}
\end{enumerate}
Output $\bfb^*$.
\end{algorithm}
\vspace{0.2cm}

A useful property of the primal-dual algorithm (and its special cases described in the sequel) is that the sparseness structure of $\bflambda_i$ conforms to the local structure of the functions $h_i$ in the following sense: assume the variables $\bfb$ are indexed by $1,...,m$, and the function $h_i(\bfb)$ depends on small subset of variables indexed by $N(i) \subset \{1,...,m\}$, then $\bflambda_{i, \alpha}^*$ contains information only for $\alpha \in N(i)$ and the remaining entries vanish.
\vspace{0.2cm}
\begin{claim}[Locality of Dual Variables]
%\label{claim:l-sparse}
Assume variables $\bfb$ are indexed by $\{1,...,m\}$ and $h_i(\bfb)$ depends only on a subset of variables indexed by $N(i) \subset \{1,...,m\}$, then the following hold: $$\bflambda \in \partial h_i(\bfb) \hspace{1cm} \Longrightarrow  \hspace{1cm}  \forall \beta \not \in N(i) \;\;\; \bflambda_\beta=0$$
\end{claim}
{\bf Proof:} Consider the decomposition of $\bfb$ to two parts $\bfb = (\bfb_{N(i)},\bfb_{\bar N(i)})$, where $\bar N(i)$ is the complement ${\{1,...,m\} \setminus N(i)}$, and likewise for the sub-gradient $\bflambda$. Following Definition \ref{def:sub-d} if $\bflambda \in \partial h(\bfb)$ then 
\be
h_i(\hat \bfb) \ge h_i(\bfb) + \bflambda^\top (\hat \bfb-\bfb)  \;\;\; \forall \hat \bfb.\label{eq:bhat}
\ee
The linear term $\bflambda^\top (\hat \bfb-\bfb)$ decomposes to the sum of $\bflambda_{N(i)}^\top (\hat \bfb_{N(i)}-\bfb_{N(i)})$ and $\bflambda_{\bar N(i)}^\top (\hat \bfb_{\bar N(i)}-\bfb_{\bar N(i)})$. Since $\hat \bfb$ is arbitrary we can choose $\hat\bfb=(\hat\bfb_{N(i)},\hat\bfb_{\bar N(i)})$ where $\hat\bfb_{N(i)}$ is set to
$\hat\bfb_{\bar N(i)}  = r (\bflambda_{\bar N(i)} - \bfb_{\bar N(i)})$ for some arbitrary scalar $r>0$, and $\hat\bfb_{\bar N(i)}$ is arbitrary. Eqn.~\ref{eq:bhat} then becomes:
$$h_i(\hat \bfb) \ge h_i(\bfb) + \bflambda_{N(i)}^\top (\hat \bfb_{N(i)}-\bfb_{N(i)}) + r \bflambda_{\bar N(i)}^\top \bflambda_{\bar N(i),}$$
for all $r>0$. If we assume to the contrary that $\bflambda_{\bar N(i)} \ne \bfzero$ then we can increase the value of $r$ and thus make the righ-hand side of the equation arbitrarily high, while not effecting the left hand side since $h_i(\hat \bfb)$ is independent of $r$ by the claim assumption (as $h_i$ depends only on the variables indexed by $N(i)$) - in contradiction to $\bflambda \in \partial h(\bfb)$.  \eop \vspace{0.2cm} 

The primal-dual algorithm is still unattractive as it requires the evaluation of the sub-differentials of $\partial f$ and $\partial h_i$ which could be as difficult as the computation of the conjugate functions. Our setting, however, is more constrained than the setting described in Theorem~\ref{theorem:dba}. In particular, the function $f=\hat f + \delta_{\cal B}$ where $\hat f$ is essentially smooth and ${\cal B}=\{\bfb\ :\ A\bfb=\bfc\}$ is an affine set. Since $f$ is non-differentiable the dual is not strictly convex, and thus we cannot expect $\bflambda_i$ to be uniquely defined. Nevertheless, we show below that $\bflambda_i$ has a convenient and simple form.
\vspace{0.2cm}
\begin{claim}
\label{claim:invariance}
Let $f(\bfb)=\hat f(\bfb)+\delta_{\cal B}(\bfb)$ where $\hat f$ is essentially smooth and ${\cal B}=\{\bfb\ :\ A\bfb=\bfc\}$ and assume that $dom(h_i)\subseteq dom(f)$. Assume 
 the functions $g(\bfb) \leftarrow  f(\bfb) + \bfb^\top \bfmu_i$ and $h(\bfb) \leftarrow h_i(\bfb)$ satisfy the assumptions of Theorem \ref{theorem:FD1}. Then for every real vector $\bfsigma$ the sub-gradient $\bflambda_i^* = -\bfmu_i - \nabla f_s(\bfb^*) + A^\top \bfsigma$ is optimal dual, i.e. satisfies Eqn. \ref{eq:pdba-opt}
\end{claim} \vspace{0.1cm}
{\bf Proof:} Theorem \ref{theorem:FD1} ensures the existence of a primal-dual pair $(\bfb^*, \bflambda^*)$ which satisfy Eqn. \ref{eq:pdba-opt}. The domains of $f(\bfb)$ and $h_i(\bfb)$ are contained in $\cal B$ by assumption, therefore by Claim \ref{claim:B} 
$$\forall \bfsigma \hspace{1cm} \bflambda^*_i + A^\top \bfsigma \in \{ -\bfmu_i - \partial f(\bfb^*) \} \cap \partial h_i(\bfb^*),$$ 
meaning that for every $\bfsigma$ the sub-gradient $(\bflambda_i^* + A^\top \bfsigma)$ is dual optimal. From linearity of the sub-differential we have $\partial f(\bfb^*) = \nabla \hat f(\bfb^*) + \partial \delta_{\cal B}$. Following Claim \ref{claim:dif-affine} the sub-differential $\delta_{\cal B}$ is represented by vectors in the linear subspace spanned by the columns of $A^\top$, denoted by $A^\top \bfsigma_0$. Using again the linearity of the sub-differential we deduce 
$$\forall \bfsigma \hspace{1cm} (\bflambda^*_i + A^\top \bfsigma)  = -\bfmu_i - \nabla \hat f(\bfb^*) + A^\top (\bfsigma-\bfsigma_0)$$ 
is a dual optimal sub-gradient. The claim follows by replacing $\bflambda_i^* \leftarrow (\bflambda^*_i + A^\top \bfsigma)$. \eop \vspace{0.2cm} 

\begin{algorithm}[Primal-Dual Ascent]
\label{alg:pdba-nonsmooth}
Let the functions $f(\bfb)$ and $h_i(\bfb)$ satisfy the conditions of Theorem~\ref{theorem:dba} where in addition  let $f(\bfb) = \hat f(\bfb) + \delta_{\cal B}(\bfb)$ where $\hat f(\bfb)$ is essentially smooth, ${\cal B}=\{\bfb\ :\ A\bfb=\bfc\}$ and $dom(h_i)\subseteq dom(f)$. Initialize $\bflambda_1=\bfzero,...,\bflambda_n=\bfzero$.
\begin{enumerate}
\item Repeat until convergence:
\item For $i=1,...n$:
  \begin{enumerate}
  \item $\bfmu_i \leftarrow   \sum_{j \ne i} \bflambda_j$
  \item $\bfb^* \leftarrow  \argminn{ \bfb \in dom(f) \cap dom(h_i)}\left\{f(\bfb) + h_i(\bfb) + \bfb^\top \bfmu_i \right\}$
  \item $\bflambda_i  \leftarrow  -\bfmu_i - \nabla \hat f(\bfb^*) + A^\top \bfsigma$
  \end{enumerate}
\end{enumerate}
Output $\bfb^*$.
\end{algorithm}
\vspace{0.2cm}

\begin{claim}[Convergence]
\label{claim:convergence}
Algorithm~\ref{alg:pdba-nonsmooth} converges to the dual and primal optimum. Moreover, its primal sequence converges to the primal optimal point $\bfb^*$ and whenever its dual sequence is bounded every of its limit point is an optimal dual solution $\bflambda_1^*,..., \bflambda_n^*$.
\end{claim} \vspace{0.1cm}
{\bf Proof:} 
Algorithm \ref{alg:pdba-nonsmooth} implicitly performs dual block ascent and the dual sequence it generates is identical to the dual sequence generated by Algorithm \ref{alg:dba} therefore inherits the features described in Theorem \ref{theorem:dba}.
Theorem \ref{theorem:FD1} relates $\bfb^*$ with the primal sequence describe in Theorem \ref{theorem:dba} by $\bfb^* = \nabla f^*(-\bflambda_i - \bfmu_i)$ \eop \vspace{0.2cm} 

The special case of Algorithm~\ref{alg:pdba-nonsmooth} when $h_i=\delta_{C_i}$, where $C_i$ is a convex set, and $f$ is essentially smooth, i.e., $A=0$, can be mapped (by eliminating step 2(a)) to a {\it successive Bregman projection\/} algorithm \cite{bregman99,censor98} which is also known under the names of Dykstra, Hildreth, Han and Csiszar. This class of iterative projection schemes has a long history starting from Von-Neumann in the 50s \cite{Von-Neumann50} who introduced the case where $f(\bfb)=\|\bfb-\bfb_0\|^2$ and $C_i$ are affine sets. In that case the primal solution is to find the projection of $\bfb_0$ onto the intersection of the affine sets $C_1\cap...\cap C_n$ and  the sub-problem in Eqn. \ref{eq:pdba} corresponds to the projection of $\bfmu_i$ onto the affine set $C_i$.
Hildreth \cite{Hildreth57} extended the problem with open half spaces $C_i=\{\bfx\ |\ \bfa_i^\top\bfx\le b_i\}$. Bregman \cite{Bregman67} extended Hildreth's problem setup by including any strictly convex function $f$. The special case of Entropy projections was introduced later by Csiszar \cite{Csiszar75}, as $I$-projections.
Dykstra \cite{Dykstra83,Dykstra85} was the first to introduce general convex sets $C_i$ (i.e., going beyond  affine sets or half-spaces) but limited the treatment to $f$ representing the Euclidean norm and the KL divergence.  The view of the algorithm with general essentially smooth $f$ and convex sets $C_i$ as performing successive Bregman projections is due to \cite{censor98,bregman99}. 

Algorithm~\ref{alg:pdba-nonsmooth} extends the body of iterative schemes mentioned above along three directions: (i) $f$ is extended to non-smooth functions which in turn makes $\bflambda_i$ non-uniquely defined, (ii) as a result $\bflambda_i$ is defined up to an additive term which in the context of the message-passing norm-product algorithm (and its special cases) translates to the normalization of the messages $n_{i\rightarrow\alpha}$, and (iii) our algorithm has two auxiliary variables, $\bfmu_i$ and $\bflambda_i$, which allows a straightforward mapping onto a message-mapping framework and complies with the local structure of the underlying graph (Claim~\ref{claim:l-sparse}).

\begin{figure*}
\fbox{
\begin{minipage}[c]{17.5cm}
\bea
\min_{b_i,b_\alpha, \alpha\in N(i)} &&  \hspace{-0.8cm} \left\{ -\sum_{x_i}b_i(x_i)\ln \phi_i(x_i)  -\sum_{\alpha\in N(i)}\sum_{\bfx_\alpha}b_\alpha(\bfx_\alpha)\ln \hat\psi_{i,\alpha}(\bfx_\alpha)
 - \epsilon\hat c_i H(\bfb_i) - \hspace{-0.2cm} \sum_{\alpha\in N(i)} \hspace{-0.2cm}  \epsilon\hat c_{i\alpha}(H(\bfb_\alpha) - H(\bfb_i))\right\}\label{eq:stepb}\\
&& subject\ to:\nonumber\\
&&  \sum_{\bfx_\alpha}b_\alpha(\bfx_\alpha) = 1,\ \sum_{\bfx_\alpha\setminus  x_i}b_\alpha(\bfx_\alpha) = b_i(x_i),\
\;\; \forall  \bfx_\alpha, \alpha\in N(i)\nonumber
\eea
\end{minipage}
}
\caption{The local sub-problem of in step (b) of Algorithm \ref{alg:pd-general} ($\min f_\epsilon(\bfb) + \bfb^\top \bfmu_i + h_{\epsilon,i}(\bfb)$) solved by the norm-product algorithm.} 
\label{fig:local-opt}
\end{figure*}

\subsection{The non-convex case}
\label{app:dykstra-nc}

So far both $f$ and $h_i$ were convex, yet  Algorithm~\ref{alg:pdba-nonsmooth} is still well defined when the functions $h_i$ are non-convex. The purpose of this section is to clarify what can be guaranteed under such conditions. We will show that indeed there is no convergence guarantees, but if the algorithm does converge then it will do so to a stationary point of the primal program.

To minimize the program $f(\bfb) + \sum_i h_i(\bfb)$ one must introduce Lagrange multipliers $\bflambda_1,...,\bflambda_n$. Whenever $f(\bfb)$ and $h_i(\bfb)$ are convex the Lagrange multipliers are the arguments of the dual function, and recovering $\bflambda_i^*$ amounts to improving the dual objective with its best $\bflambda_i$-arguments, therefore this procedure is guaranteed to converge. When $h_i$ are non-convex  the Lagrange multipliers {\em do not} correspond to a dual function, and thus recovering $\bflambda_i^*$ amounts to finding a stationary point with respect to a sub-problem involving $f(\bfb)$ and a single $h_i(\bfb)$, and convergence cannot be guaranteed in general. Nevertheless, in each iteration we recover Lagrange multipliers for a stationary point of a related sub-problem, therefore, intuitively, {\em if} this method converges, it reaches a stationary point of the non-convex program $f(\bfb) + \sum_i h_i(\bfb)$.

We consider programs with non-convex smooth functions $h_i(\bfb)$ restricted  to the affine domain $\{\bfb: A_i \bfb = \bfc_i\}$, and Legendre-type function $f(\bfb)$, whose conjugate function is finite. Recall Theorem \ref{theorem:smooth}, describing Legendre-type function as an essentially smooth function which is strictly convex in its interior and  satisfies $\nabla f^* = (\nabla f)^{-1}$. For this type of non-convex programs we show in the following claim, that if  Algorithm~\ref{alg:pdba-nonsmooth}  converges, it reaches a local-minimum of $f(\bfb) + \sum_i h_i(\bfb)$: \vspace{0.2cm}
\begin{claim}
Consider Algorithm~\ref{alg:pdba-nonsmooth}  with $A=0$ for Legendre-type function $f(\bfb)$  and non-convex continuously differentiable functions $h_i(\bfb)$ restricted to the affine domain $\{\bfb: A_i \bfb = \bfc_i\}$, and assume $\bfb^*$ in Eqn. \ref{eq:pdba} is in the interior of $dom(f)$ relative to the affine set $dom(h_i)$. Then if the algorithm converges it reaches a stationary point of the non-convex program $f(\bfb) + \sum_i h_i(\bfb)$.
\end{claim}\vspace{0.1cm}
{\bf Proof:}
The optimization $$ \bfb^{*,(i)} = \argminn{\bfb \in dom(h_i)} \left\{ f(\bfb) + \bfb^\top \bfmu_i + h_i(\bfb) \right\} $$ satisfies the conditions of the Lagrange multiplier Theorem \ref{theorem:Lagrange-d} with respect to the affine set $dom(h_i)$, therefore if algorithm converges there holds $$(*) \; \forall i \;\;\;\;\;\; \nabla f(\bfb^{*,(i)}) + \bfmu_i + \nabla h_i(\bfb^{*,(i)}) + A_i^\top \bfnu_i^* = 0$$

From steps 2a and 2c, for every $i$ there must hold $\sum_{j=1}^n \bflambda_j = -\nabla f(\bfb^{*,(i)})$. The conjugate of Legendre-type function satisfies $\nabla f^* = (\nabla f)^{-1}$ by Theorem \ref{theorem:smooth}, therefore for every $i$ holds $\bfb^{*,(i)} = \nabla f^*(-\sum_j \bflambda_i)$. This implies that the local primal arguments $\bfb^{*,(i)}$ are the same for every $i$, and we denote them by $\bfb^*$.  Summing up the relations in (*) we get $$\sum_i \nabla h_i(\bfb^*) + n \nabla f(\bfb^*) + \sum_{i=1}^n \bfmu_i + \sum_{i=1}^n A_i \bfnu_i^*= 0.$$
Substituting $\bfmu_i = -\bflambda_i - \nabla f(\bfb^*)$ (from step 2c) we obtain  the stationary condition for $\bfb^*$, i.e. $\nabla f(\bfb^*) + \sum_i \nabla h_i (\bfb^*) + \sum_{i=1}^n A_i \bfnu_i^* = 0$. \eop \vspace{0.2cm} 

\subsection{The non-strictly convex case}
\label{app:non-strict}

The case $\epsilon=0$ in eqn.~\ref{eq:LPe} corresponds to having a non-strictly convex function $f_\epsilon$ in eqn.~\ref{eq:f}. This situation can be analyzed in greater generality by observing the behavior of Algorithm \ref{alg:pdba} when the function $f$ is convex but not strictly convex. 

For convex $f(\bfb)$ and $h_i(\bfb)$ the primal program in Eqn. \ref{eq:primal} upper bounds the dual function in Eqn. \ref{eq:dual}, and the dual block ascent optimization scheme which iteratively improves the dual function must converge. If the function $f(\bfb)$ is not strictly convex its conjugate is not smooth and the dual block ascent is not guaranteed to reach the global optimum. We describe, in a nutshell, where things go wrong in the Algorithm \ref{alg:pdba}: Assume the algorithm converges, then for every $i$ we obtain the primal solution $$   \bfb^{*,(i)} = \argminn{\bfb \in dom(f) \cap dom(h_i)} \left\{ f(\bfb) + \bfb^\top \bfmu^*_i + h_i(\bfb) \right\}$$ Recovering the dual variables corresponds to finding $\bflambda_i^* \in \{-\bfmu^*_i - \partial f(\bfb^{*,(i)})\}$. Recall that $\bfmu^*_i = \sum_{j \ne i} \bflambda^*_j$, then the primal-dual relation boils down to $-\sum_j \bflambda^*_j  = \partial f(\bfb^{*,(i)})$ for every $i$. If $f$ was strictly convex it would have imply that all the $\bfb^{*,(i)}$ are in fact the same $\bfb^*$ and it would have ensure optimality. Since $f(\bfb)$ is not strictly convex it means that the algorithm might converge in the dual domain but we cannot recover a consistent $\bfb^*$.

\section{The Norm-Product Algorithm}
\label{app:np}

\begin{figure*}
\fbox{
\begin{minipage}[c]{18cm}
\be
\min_{b_i(x_i)\in{\cal P}}\hspace{-0.1cm} \left\{ \hspace{-0.15cm}  -\sum_{x_i} b_i(x_i) \ln \phi_i(x_i)-\epsilon\hat c_i H(\bfb_i) + \hspace{-0.1cm} \sum_{x_i} b_i(x_i) \hspace{-0.2cm} \underbrace{\sum_{\alpha\in N(i)} \hspace{-0.2cm} \epsilon\hat c_{i\alpha}\left[\underbrace{\min_{b_{\alpha | i} \in{\cal P}}-\sum_{\bfx_\alpha\setminus  x_i}\bai(\xai)\ln\hat\psi_{i,\alpha}^{1/(\epsilon\hat c_{i\alpha})}(\bfx_\alpha) - H(\bai)}_{(*)}\right]}_{(**)}\right\}\label{eq:stepb2}
\ee
\end{minipage}
}
\caption{Reducing the local sub-problem in Fig. \ref{fig:local-opt} to a series of normalizations by introducing conditional entropies.}
\label{fig:local-sol}
\end{figure*}

We embed the function definitions of $f_\epsilon$ and $h_{\epsilon,i}$ into the primal-dual Algorithm~\ref{alg:pd-general}. 
Given the sparse structure of $h_{\epsilon,i}$  then, following Claim~\ref{claim:l-sparse}, we present the entries of $\bflambda_i$ according to the factor-graph structure by setting  $\bflambda_{i}=\{\lambda_{i,\alpha}(\bfx_\alpha)\}$ (and likewise $\bfmu_{i,\alpha}$). We first define few short-cut notations:
\bea
\hat\psi_{i,\alpha}(\bfx_\alpha) &\stackrel{def}{=}& \psi_\alpha(\bfx_\alpha)\exp(-\mu_{i,\alpha}(\bfx_\alpha)),\label{eq:psihat}\\
 \hat c_{i\alpha} &\stackrel{def}{=}& c_\alpha + c_{i\alpha},\nonumber\\
 \hat c_i &\stackrel{def}{=}& c_i + \sum_{\alpha\in N(i)} c_\alpha\nonumber
\eea
Step (b) of Algorithm~\ref{alg:pd-general} is reduced to finding $\bfb^*_\alpha$ for all $\alpha\in N(i)$, described in eqn.~\ref{eq:stepb} in Fig. \ref{fig:local-opt}:
 
We will derive the optimal $\bfb^*_\alpha$ and show it has a closed-form solution. In the process we will be relying on the following observation which we present as a Lemma, without a proof: \vspace{0.2cm} 
\begin{lemma}
\label{lemma1}
Let $\psi$ be a non-negative array and $\bfp^*$ be the  optimal probability array for the following optimization problem:
$$\bfp^*=\argminn{p(\bfx) \ge 0, \sum_{\bfx} p(\bfx)=1}\left\{- \sum_{\bfx} p(\bfx)\ln\psi(\bfx) - H(\bfp)\right\},$$
then,
\bea
p^*(\bfx)&=&\frac{1}{\sum_{\bfy} \psi(\bfy)}\psi(\bfx) \label{eq:lemma11}\\
-\ln \sum_{\bfx} \psi(\bfx)&=& - \sum_{\bfx} p^*(\bfx)\ln\psi(\bfx) - H(\bfp^*)\ \ \  \ \label{eq:lemma12}
\eea
\end{lemma} \eop \vspace{0.3cm} 

We will be repeatedly using Lemma~\ref{lemma1} in the derivation of $\bfb^*_\alpha$, as follows. Let $\bai(\xai)$ and $H(\bai)$ be defined below:
\beas
\bai(\xai) &\stackrel{def}{=}& \frac{b_\alpha(\bfx_\alpha)}{b_i(x_i)}\\
H(\bai)&\stackrel{def}{=}& - \sum_{\bfx_\alpha\setminus  x_i} \frac{b_\alpha(\bfx_\alpha)}{b_i(x_i)}\ln \frac{b_\alpha(\bfx_\alpha)}{b_i(x_i)}. 
%= - \sum_{\bfx_\alpha\setminus  x_i}\bai(\xai)\ln\bai(\xai)
\eeas
Note that the constraint $\bai(\xai)\in{\cal P}$, i.e., that $\bai$ lives in the probability simplex, is equivalent to the marginal consistency constraint $\sum_{\bfx_\alpha\setminus x_i}b_\alpha(\bfx_\alpha)=\bfb(x_i)$ as well. We can use $H(\bai)$ to simplify  the conditional entropy term $H(\bfb_\alpha) - H(\bfb_i)$ by the following Lemma: \vspace{0.2cm} 
\begin{lemma}
$$H(\bfb_\alpha) - H(\bfb_i)=\sum_{ x_i} b_i(x_i)H(\bai)$$
\end{lemma} \vspace{0.1cm}
{\bf Proof:\ } 
The Lemma is based on the definition of conditional entropy $H(X\cond Y)=H(X,Y)-H(Y)=\sum_{y}p(y)H(X\cond Y=y)$ for random variables $X,Y$. In our terms we have $H(\bfx_\alpha\setminus x_i \cond  x_i)=H(\bfb_\alpha)-H(\bfb_i)=\sum_{ x_i}b_i(x_i)H(\bai)$.
 \eop \vspace{0.2cm} 

With the definitions above, the optimization problem of step (b) as described in eqn.~\ref{eq:stepb}, can be broken down to a cascade of two steps, described in eqn.~\ref{eq:stepb2} in Fig. \ref{fig:local-sol}.

From Lemma~\ref{lemma1} (eqn.~\ref{eq:lemma12}) we obtain the solution for the inner optimization block $(*)$:
$$(*) = -\ln  \sum_{\bfx_\alpha\setminus  x_i}\hat\psi_{i,\alpha}(\bfx_\alpha)^{1/(\epsilon\hat c_{i\alpha})}.$$

We make the following definition:
\bea
m_{\alpha\rightarrow i}(x_i)&\stackrel{def}{=}&\left(\sum_{\bfx_\alpha\setminus  x_i}\hat\psi_{i,\alpha}(\bfx_\alpha)^{1/(\epsilon\hat c_{i\alpha})}\right)^{\epsilon\hat c_{i\alpha}}\nonumber\\
\label{eq:mai}
\eea
Therefore, the inner-block denoted by $(**)$ takes the form:
$$(**) = -\ln\prod_{\alpha\in N(i)}m_{\alpha\rightarrow i}(x_i).$$
Substituting  $(**)$ back into eqn.~\ref{eq:stepb2} we obtain:
{\small $$
\min_{b_i(x_i)\in{\cal P}}\epsilon\hat c_i  \hspace{-0.1cm} \left[ -H(\bfb_i) - \sum_{x_i}b_i(x_i) \ln \phi^{1/\epsilon\hat c_i}_i(x_i) \hspace{-0.15cm} \prod_{\alpha\in N(i)}  \hspace{-0.15cm} m^{1/\epsilon\hat c_i}_{\alpha\rightarrow i}(x_i)\right]
$$}
and from Lemma~\ref{lemma1} (eqn.~\ref{eq:lemma11}) we obtain a closed-form solution for $b^*_i(x_i)$:
\be
b^*_i(x_i)\propto \left(\phi_i(x_i) \prod_{\alpha\in N(i)}m_{\alpha\rightarrow i}(x_i)\right)^{1/\epsilon\hat c_i}.\label{eq:bi}
\ee
Finally, $b^*_\alpha(\bfx_\alpha)=\bai^*(\xai)b^*_i(x_i)$ takes the form:
\be
b^*_\alpha(\bfx_\alpha)= \frac{b^*_i(x_i)}{m_{\alpha\rightarrow i}^{1/\epsilon\hat c_{i\alpha}}(x_i)}\hat\psi_{i,\alpha}(\bfx_\alpha)^{1/(\epsilon\hat c_{i\alpha})}\label{eq:ba}
\ee

Next we evaluate step (c) of Algorithm~\ref{alg:pd-general}, i.e.,
\be
\bflambda^*_{i,\alpha}(\bfx_\alpha) =  -\mu_{i,\alpha}(\bfx_\alpha) - \nabla \hat f_\epsilon(b^*_\alpha(\bfx_\alpha)) + \sigma_\alpha\bfone,\label{eq:la}
\ee
where $\sigma_\alpha$ is an arbitrary scalar and $\hat f_\epsilon$ is defined in eqn.~\ref{eq:f}. Define $n_{i\rightarrow\alpha}(\bfx_\alpha)$ as follows:
\be
n_{i\rightarrow\alpha}(\bfx_\alpha)\stackrel{def}{=} \exp(-\lambda_{i,\alpha}(\bfx_\alpha))
\ee
We note, therefore, that the additive constant freedom in the definition of $\bflambda_{i,\alpha}$ becomes a scaling choice in the definition of $n_{i\rightarrow\alpha}$. Without loss of generality we choose the scale such that $n_{i\rightarrow\alpha}(\bfx_\alpha)\in{\cal P}$. The claim below sets the value of $n_{i\rightarrow\alpha}$: \\
\begin{proposition}
\be
n_{i\rightarrow\alpha}(\bfx_\alpha)\propto \left(\frac{b^*_i(x_i)}{m_{\alpha\rightarrow i}^{1/\epsilon\hat c_{i\alpha}}(x_i)}\right)^{\epsilon c_\alpha}\hat\psi_{i,\alpha}(\bfx_\alpha)^{-c_{i\alpha}/\hat c_{i\alpha}}\label{eq:nia1}
\ee
\end{proposition}
{\bf Proof:\ }
From definition of $n_{i\rightarrow\alpha}$ and from eqn.~\ref{eq:la} we have:
$$n_{i\rightarrow\alpha}(\bfx_\alpha)\propto \exp(\mu_{i,\alpha}(\bfx_\alpha))\exp( \nabla \hat f_\epsilon(b^*_\alpha(\bfx_\alpha))).$$
Substituting the value of $ \nabla \hat f_\epsilon(b^*_\alpha(\bfx_\alpha))$:
$$ \nabla \hat f_\epsilon(b^*_\alpha(\bfx_\alpha))=-\ln\psi_\alpha(\bfx_\alpha) + \epsilon c_\alpha(\ln b^*_\alpha(\bfx_\alpha) + 1),$$
and the value of $b^*_\alpha(\bfx_\alpha)$ from eqn.~\ref{eq:ba} we obtain:
\beas
n_{i\rightarrow\alpha}(\bfx_\alpha)&\propto&\exp(\mu_{i,\alpha}(\bfx_\alpha)-\ln\psi_\alpha(\bfx_\alpha))(b^*_\alpha(\bfx_\alpha))^{\epsilon c_\alpha}\\
&=&\hat\psi^{-1}_{i,\alpha}(\bfx_\alpha)\left(\frac{b^*_i(x_i)}{m_{\alpha\rightarrow i}^{1/\epsilon\hat c_{i\alpha}}(x_i)}\right)^{\epsilon c_\alpha}\hat\psi^{c_\alpha/\hat c_{i\alpha}}_{i,\alpha}(\bfx_\alpha)\\
&=&\left(\frac{b^*_i(x_i)}{m_{\alpha\rightarrow i}^{1/\epsilon\hat c_{i\alpha}}(x_i)}\right)^{\epsilon c_\alpha}\hat\psi^{c_\alpha/\hat c_{i\alpha}-1}_{i,\alpha}(\bfx_\alpha)
\eeas
and following substitution of $\hat c_{i\alpha}=c_\alpha + c_{i\alpha}$ we obtain what we set out to prove.
\eop \vspace{0.2cm} 

Substituting $\bfmu_{i,\alpha}=\sum_{j\in N(\alpha)\setminus i} \bflambda_{j,\alpha}$ into the definition of $\hat\psi_{i,\alpha}$ (eqn.~\ref{eq:psihat}) we obtain:
\bea
\hat\psi_{i,\alpha}(\bfx_\alpha) &=& \psi_\alpha(\bfx_\alpha)\prod_{j\in N(\alpha)\setminus i}\exp(-\bflambda_{j,\alpha}(\bfx_\alpha))\nonumber\\
&=& \psi_\alpha(\bfx_\alpha)\prod_{j\in N(\alpha)\setminus i}n_{j\rightarrow\alpha}(\bfx_\alpha)\label{eq:psihat2}
\eea
Substituting eqn.~\ref{eq:psihat2} into eqn.~\ref{eq:mai} we obtain the update rule for $m_{\alpha\rightarrow i}$:
\be
m_{\alpha\rightarrow i}(x_i)= \left( \sum_{ \bfx_\alpha\setminus x_i} \left( \psi_\alpha(\bfx_\alpha)\prod_{j\in N(\alpha)\setminus i}n_{j\rightarrow\alpha}(\bfx_\alpha) \right)^{1/\epsilon\hat c_{i\alpha} } \right)^{\epsilon\hat c_{i\alpha}}.
\label{eq:mai2}
\ee
Substituting eqn.~\ref{eq:psihat2} into eqn.~\ref{eq:nia1} we obtain:

{\small $$
n_{i\rightarrow\alpha}(\bfx_\alpha)\propto \left(\frac{b^*_i(x_i)}{m_{\alpha\rightarrow i}^{1/\epsilon\hat c_{i\alpha}}(x_i)}\right)^{\epsilon c_\alpha}\left(\psi_\alpha(\bfx_\alpha)  \hspace{-0.2cm} \prod_{j\in N(\alpha)\setminus i} \hspace{-0.2cm} n_{j\rightarrow\alpha}(\bfx_\alpha)\right)^{\frac{-c_{i\alpha}}{\hat c_{i\alpha}}},
$$
}

and substituting $\bfb_i^*$ in eqn. \ref{eq:bi} we obtain the update rule for $n_{i\rightarrow\alpha}$. 
%&\propto&  \left(\frac{\displaystyle \phi_i^{1/\hat c_i}(x_i) \prod_{\beta\in N(i)} m_{\beta\rightarrow i}^{1/\hat c_i}(x_i)}{m_{\alpha\rightarrow i}^{1/\hat c_{i\alpha}}(x_i)}\right)^{c_\alpha}\left(\psi_\alpha(\bfx_\alpha)\prod_{j\in N(\alpha)\setminus i}n_{j\rightarrow\alpha}(\bfx_\alpha)\right)^{-c_{i\alpha}/\hat c_{i\alpha}}\label{eq:nia}

%\bea
%n_{i\rightarrow\alpha}(\bfx_\alpha)&\propto& \left(\frac{b^*_i(x_i)}{m_{\alpha\rightarrow i}^{1/\epsilon\hat c_{i\alpha}}(x_i)}\right)^{\epsilon c_\alpha}\left(\psi_\alpha(\bfx_\alpha)\prod_{j\in N(\alpha)\setminus i}n_{j\rightarrow\alpha}(\bfx_\alpha)\right)^{-c_{i\alpha}/\hat c_{i\alpha}}\nonumber\\
%&\propto&  \left(\frac{\displaystyle \phi_i^{1/\hat c_i}(x_i) \prod_{\beta\in N(i)} m_{\beta\rightarrow i}^{1/\hat c_i}(x_i)}{m_{\alpha\rightarrow i}^{1/\hat c_{i\alpha}}(x_i)}\right)^{c_\alpha}\left(\psi_\alpha(\bfx_\alpha)\prod_{j\in N(\alpha)\setminus i}n_{j\rightarrow\alpha}(\bfx_\alpha)\right)^{-c_{i\alpha}/\hat c_{i\alpha}}\label{eq:nia}
%\eea

%\section{The setting of Parameters Corresponding to the Tree-reweighted (TRW) free-energy}
%\label{app:trw}

\section{Convex-Free-Energy Parameter Settings}
\label{app:confree}

The fractional entropy approximation of eqn.~\ref{eq:Hcon} 
$$ \sum_\alpha \bar c_\alpha H(\bfb_\alpha) + \sum_i \bar c_i H(\bfb_i),$$
is strictly convex if it can be written as eqn.~\ref{eq:H} 
$$\sum_\alpha c_\alpha H(\bfb_\alpha) + \sum_i c_i H(\bfb_i) + \sum_{i,\alpha\in N(i)} c_{i \alpha} (H(\bfb_\alpha) - H(\bfb_i)),$$
in terms of $c_\alpha>0,c_i,c_{i\alpha}\ge 0$. In this section we will introduce a number entropy approximations which fall into the convex-free-energy class. We will start with the Tree-re-weighted (TRW) entropy approximation \cite{Wainwright-nips02} and then introduce other approximations.

There are two ways, introduced in the literature so far, to set parameters for the TRW entropy approximation --- both of which do not belong the required setup of a convex-free-energy. In the first version,  the TRW-free-energy corresponds to the setting of $c_\alpha>0, c_i=1-\sum_{\alpha \in N(i)} c_\alpha$ and $c_{i\alpha}=0$, where the setting of $c_\alpha$ corresponds to the relative number of spanning trees (or hyper-trees) of the graph which include the edge (hyperedge) $\alpha$. The problem with this setting is that $c_i<0$, thus, even though the fractional entropy approximation is convex, the functions  $h_i$ (defined in terms of $c_i$ and $c_{i\alpha}$) are not convex.

The second version, introduced by \cite{Globerson-UAI07}, sets  $c_i$ as the relative number of spanning trees that have node $i$ as a root,  and for an edge $\alpha = (i,j)$, $c_{i \alpha}$ is the relative number of trees that include the directed edge $j \rightarrow i$. It is possible to find such edge probabilities for the uniform distribution over all spanning trees by employing a variant of the matrix tree theorem for directed trees, \cite{Globerson-UAI07}, \cite{Tutte01} p.141. In this formulation $c_i, c_{i \alpha} > 0$ but $c_\alpha = 0$. The problem with $c_\alpha=0$ is that the function $f$ is no longer strictly convex.

In the claim below we show how to convert a TRW setting according to the second version, i.e., where $c_i, c_{i \alpha} > 0,c_\alpha=0$ to the convex-free-energy setting $c'_\alpha>0,c'_i,c'_{i\alpha}\ge 0$: \vspace{0.2cm}
\begin{claim}
Assume an approximated entropy $\sum_\alpha \bar c_\alpha H(\bfb_\alpha) + \sum_i \bar c_i H(\bfb_i)$ is described by $c_i,c_{i \alpha}>0$ and $c_\alpha =0$, i.e. $\bar c_\alpha = c_\alpha + \sum_{i \in N(\alpha)} c_{i \alpha}$ and $\bar c_i = c_i - \sum_{\alpha \in N(i)} c_{i \alpha}$. Then there exists $c_\alpha',c_i',c_{i \alpha}' >0$ which agree on the approximated entropy, namely $\bar c_\alpha = c_\alpha' + \sum_{i \in N(\alpha)} c_{i \alpha}'$ and $\bar c_i = c_i' - \sum_{\alpha \in N(i)} c_{i \alpha}'$
\end{claim} \vspace{0.1cm}
{\bf Proof:} We describe an efficient algorithm for constructing the desired convex free energy: Initialize $c_\alpha'=0, c_i'=0, c_{i \alpha}'=0$. For every $i=1,...,n$ and every $\alpha \in N(i)$ consider the entropy combination $$c_{i \alpha} (H(\bfb_\alpha) - H(\bfb_i)) + \frac{c_i}{d_i} H(\bfb_i)$$ and divide it to two cases:
\begin{enumerate}
\item $c_{i \alpha} \le c_i/d_i$, then the entropy can be equivalently written by the entropy $$c_{i \alpha} H(\bfb_\alpha) + \left(\frac{c_i}{d_i} - c_{i \alpha} \right) H(\bfb_i)$$
therefore perform
\begin{enumerate}
\item $c_\alpha' \leftarrow c_\alpha' + c_{i \alpha}$
\item $c_i' \leftarrow c_i' + (\frac{c_i}{d_i} - c_{i \alpha})$.
\end{enumerate}
\item $c_{i \alpha} > c_i/d_i$, then the entropy can be equivalently presented as:
 $$ \frac{c_i}{d_i} H(\bfb_\alpha) + \left( c_{i \alpha} - \frac{c_i}{d_i} \right) (H(\bfb_\alpha) - H(\bfb_i)),$$
therefore perform
\begin{enumerate}
\item $c_\alpha' \leftarrow c_\alpha' + \frac{c_i}{d_i}$
\item $c_{i \alpha}' \leftarrow c_{i \alpha}' + (c_{i \alpha} - \frac{c_i}{d_i})$.
\end{enumerate}
Since $c_i,c_{i \alpha}$ are positive we obtain an equivalent entropy approximation with $c_\alpha'>0$ and $c_i', c_{i \alpha}'\ge 0$. A straight forward bookkeeping ensures that $\bar c_\alpha$ and $\bar c_i$ do not change.
\end{enumerate}
\eop \vspace{0.2cm} 

Another concave approximation is to seek a setting of parameters $c_\alpha>0,c_i,c_{i\alpha}\ge 0$ such that the approximation $\tilde H$ is as close as possible to the non-convex Bethe approximation\footnote{A similar idea was  independently derived by Nir Friedman and his collaborators --- personal communication.}. Given the equations in Definition~\ref{def:cfe} connecting the parameters $c_\alpha,c_i,c_{i\alpha}$ to $\bar c_\alpha$ and $\bar c_i$, the space of admissible solutions must satisfy the following equations:
\beas
&&c_i + \sum_{\alpha \in N(i)} ( c_{\alpha} + \sum_{j \in N(\alpha) \setminus i} c_{j \alpha} ) = 1,\ \ i=1,...,n\\
&&c_i,c_{i \alpha} \ge 0,\ \ c_{\alpha} > 0.
\eeas
Among all possible admissible solutions we choose the one in which $\bar c_\alpha$ is as uniform as possible, i.e., we apply Laplace's principle of insufficient reasoning. The criterion function, therefore,  minimizes:
\be \min_{c_i,c_{i \alpha},c_\alpha \in admissible} \sum_{\alpha} (c_{\alpha} + \sum_{i \in N(\alpha)} c_{i\alpha}  - 1)^2,\label{eq:maxent}
\ee
which is a least-squares criteria for uniformity of $\bar c_\alpha$. We refer to the two least-squares scheme as $L_2$ convex free energy approximation. In an earlier work \cite{Hazan-Shashua-uai08}, we also used the maximum entropy approach where the criterion function minimizes $\sum_\alpha \bar c_\alpha\ln \bar c_\alpha$. Further investigation for constructing good convex free energy approximations can be found in \cite{Meshi-uai09}.  

The desire towards uniformity, besides being used extensively in probabilistic settings, is motivated by the success of the Bethe free energy where $\bar c_\alpha = 1$. The Bethe free energy is non-convex for factor graphs with cycles, thus is not a member of the convex free energies, but empirical evidence suggest that when BP converges the marginals are surprisingly good. For Bethe free energy $\bar c_\alpha = 1$ over all factor nodes $\alpha$ --- hence our proposal to strive for uniformity over the space of admissible solutions. In some sense we are attempting to "convexify" the Bethe free energy, although this is not being done directly. 

\section{Incorporating zero potentials}

A particularly important class of factors are those with zero potentials. These type of potentials are used, for example, in defining error-correcting codes.  Note that if one or more of the factor potentials $\psi_\alpha(x_\alpha)$ or local potentials $\phi(x_i)$ are equal to zero, then the overall probability of states which contain these configurations is zero, namely $p(\bfx_\alpha)=0$ or $p(x_i)=0$ respectively. This restriction on the marginal probabilities implicitly appears in the variational programs using the convention $0 \ln 0 = 0$ and $-x \ln 0 = \infty$ for $x>0$. 

Recall that the variational approach seeks a distribution $p(x_1,...,x_n)$ which is as close as possible, in relative entropy terms, to the product $\prod_i \phi_i(x_i) \prod_\alpha \psi_\alpha(\bfx_\alpha)$. Expanding the relative entropy produces the free energy: $$g(\bfp) = \sum_{\alpha,\bfx_\alpha} \theta_\alpha(\bfx_\alpha) p(\bfx_\alpha) + \sum_{i, x_i}\theta_i (x_i) p(x_i) - H(\bfp),$$
where $\theta_\alpha = -\ln \psi_\alpha$, $\theta_i = -\ln \phi_i$, and $p(\bfx_\alpha)$, $p(x_i)$ are the marginal probabilities, and $H(\bfp)$ is the entropy function. Since $-x \ln 0 = \infty$ whenever $x>0$, the zero potential $\psi_\alpha(\bfx_\alpha)=0$ constraints $\bfp \in dom(g)$ if and only if $p(\bfx_\alpha)=0$, Likewise, $\phi_i(x_i)=0$ constrains $\bfp \in dom(g)$ whenever $p(x_i)=0$. Following the above, the inference program which corresponds to the free energy minimization is well defined for zero potentials when we consider the domain of the free energy, and takes the form $\min_{\bfp \in dom(g)} g(\bfp)$. In spite the mathematical difficulty introduced by using zero potentials, it makes no difference from algorithmic perspective, since the optimal distribution $\bfp^*$ is the normalization of the product of potentials, $\bfp^* \propto \prod_i \phi_i \prod_\alpha \psi_\alpha$ and it respects the domain constraint. 

The same behavior appears in the variational approach for MAP assignment, where one seeks vector $\bfx^*$ which maximize the energy $\prod_i \phi_i (x_i) \prod_\alpha \psi_\alpha(\bfx_\alpha)$. This task is described by the linear function $$g(\bfp) = \sum_{\alpha,x_\alpha} \theta_\alpha(\bfx_\alpha) p(\bfx_\alpha) + \sum_{i, x_i}\theta_i (x_i) p(x_i),$$ whereas the zero potentials of the form $\phi_i(x_i)=0$ or $\psi_\alpha(\bfx_\alpha)=0$ constraints $\bfp \in dom(g)$ if and only if $p(x_i)=0$ or $p(\bfx_\alpha)=0$ respectively. Again, this mathematical nuance makes no difference from algorithmic perspective, since the optimal distribution $\bfp^*$ is the a zero-one distribution, i.e. $p^*(\bfx^*)=1$ and for every $\bfx \ne \bfx^*$ holds $p^*(\bfx)=0$.

The framework of incorporating zero potentials in the domain of the optimized program also corresponds to the approximate inference and LP-relazation described by the minimization of the function 
$$g(\bfb) = \sum_{i,x_i} \theta_i(x_i)b_i(x_i) + \sum_{\alpha,\bfx_\alpha} \theta_\alpha(\bfx_\alpha)b_\alpha(\bfx_\alpha) - \epsilon\tilde H(\bfb),$$
whose domain is constrained by zero potentials, namely $\phi_i(x_i)=0$ or $\psi_\alpha(\bfx_\alpha)=0$ constraints $\bfb \in dom(g)$ if and only if $b_i(x_i)=0$ or $b_\alpha(\bfx_\alpha)=0$ respectively. The domain constrains are inherited by the norm-product algorithm where we represent $g(\bfb)$ in the form $f_\epsilon(\bfb)+\sum_i h_{\epsilon,i}(\bfb)$ described in eqns. \ref{eq:f}, \ref{eq:hi}. In particular, $\bfb \in dom(f_\epsilon)$ only if $b_\alpha(\bfx_\alpha)=0$ whenever $\psi_\alpha(\bfx_\alpha)=0$, and $\bfb \in dom(h_{\epsilon,i})$ only if $b_i(x_i)=0$ whenever $\phi_i(x_i)=0$. This domain constraint do not affect the norm-product algorithm whose optimal beliefs are a (power) normalization of the potentials multiplied by the messages, described in eqn.~\ref{eq:bi} and eqn.~\ref{eq:ba}. Therefore, in the norm-product optimization framework, zero potential $\psi_\alpha(\bfx_\alpha)=0$ or $\phi_i(x_i)=0$ induces optimal beliefs satisfying $b_\alpha^*(\bfx_\alpha)=0$ or $b_i^*(x_i)=0$ respectively.

\section{Acknowledgements}
The authors wish to Chen Yanover, Talya Meltzer, Ofer Meshi, Ariel Jaimovich, Amir Globerson and Yair Weiss for helpful discussions. We also thank the anonymous referees for their criticisms and suggestions.

\bibliographystyle{IEEEtranS}

\bibliography{norm-prod-ieee-final}

%% biography section
%% 
%% If you have an EPS/PDF photo (graphicx package needed) extra braces are
%% needed around the contents of the optional argument to biography to prevent
%% the LaTeX parser from getting confused when it sees the complicated
%% \includegraphics command within an optional argument. (You could create
%% your own custom macro containing the \includegraphics command to make things
%% simpler here.)
%%\begin{biography}[{\includegraphics[width=1in,height=1.25in,clip,keepaspectratio]{mshell}}]{Michael Shell}
%% or if you just want to reserve a space for a photo:

%\end{IEEEbiography}


%% insert where needed to balance the two columns on the last page with
%%\newpage


%% You can push biographies down or up by placing
%% a \vfill before or after them. The appropriate
%% use of \vfill depends on what kind of text is
%% on the last page and whether or not the columns
%% are being equalized.

%%\vfill

%% Can be used to pull up biographies so that the bottom of the last one
%% is flush with the other column.
%%\enlargethispage{-5in}

\end{document}